%% file: main.tex
\documentclass[11pt,onecolumn]{IEEEtran}
\IEEEoverridecommandlockouts

\usepackage[T1]{fontenc}
\usepackage{cite}
\usepackage{enumitem}
\usepackage{amsmath,amssymb,amsfonts}
\usepackage{graphicx}
\usepackage{textcomp}
\usepackage{xcolor}
\usepackage{tikz}
\usepackage{adjustbox}
\usepackage{url}
\usepackage{booktabs}
\usepackage{array}
\usepackage{longtable}
\usepackage{float}
\usepackage{microtype}
\usepackage[hidelinks]{hyperref}
\hypersetup{
  pdftitle={From Cognitive Architectures to Language Agents: A Mechanism-Level Review of Lineage, Convergence, and Migration Gaps},
  pdfauthor={Haodi Fan and Zucong Lan}
}
\usetikzlibrary{shapes.geometric,shapes.misc,arrows,arrows.meta,positioning,fit,backgrounds,calc,decorations.pathreplacing,matrix}

\definecolor{accentblue}{RGB}{70,130,180}
\definecolor{accentgreen}{RGB}{50,150,50}
\definecolor{accentorange}{RGB}{230,130,20}
\definecolor{accentred}{RGB}{185,55,55}

\makeatletter
\let\cogarch@thebibliography\thebibliography
\renewcommand{\thebibliography}[1]{%
  \cogarch@thebibliography{#1}%
  \fontsize{7.4pt}{7.7pt}\selectfont
  \setlength{\itemsep}{0pt}%
  \setlength{\parskip}{0pt}%
}
\makeatother

\title{From Cognitive Architectures to Language Agents:\\
A Mechanism-Level Review of Lineage, Convergence, and Migration Gaps}

\author{Haodi Fan \quad Zucong Lan\\[2pt]
\href{mailto:anthonyfan@metainflow.cn}{anthonyfan@metainflow.cn}
\quad
\href{mailto:neillan@metainflow.cn}{neillan@metainflow.cn}}

\begin{document}
\maketitle

\begin{abstract}
Memory, planning, reflection, and tool use are often compared as feature labels, obscuring the control semantics that determine how an agent actually runs. This review connects ten historical cognitive architectures, eight language-agent runtime families, and forty-two mechanism-focused modern systems. We reconstruct each mechanism through state, control, transition, persistence, failure, learning, and resource governance, then code evidence relation (E1--E4) separately from migration depth (D0--D4). The resulting landscape is uneven. Modern agents have operationalized substantial parts of adaptive memory, failure recovery, dynamic team selection, workflow search, skill induction, resource scheduling, and uncertainty-conditioned action, although often through independent convergence rather than documented inheritance. The strongest remaining opportunities lie in couplings among mechanisms. Closest-baseline screening closes one proposed gap: GraSP already combines calibrated multi-skill selection, typed compilation, verification, bounded repair, and replanning or ReAct fallback. Five residual bundles remain: activation with latency and action utility; typed impasse with isolated substates and resolution compilation; bounded content competition with broadcast and admission learning; persistent intention with reconsideration and live method authority; and uncertainty with resource allocation, interruption, and stopping. We contribute a distinctive-mechanism catalog, an auditable evidence--depth framework, and a falsifiable agenda for testing these bundles as composable runtime invariants.
\end{abstract}

\input{chapters/01_introduction}
\input{chapters/02_review-method}
\input{chapters/03_historical-architectures}
\input{chapters/04_modern-runtimes}
\input{chapters/05_pairwise-structural-mapping}
\input{chapters/06_migration-agenda}
\input{chapters/07_discussion-conclusion}
\input{chapters/08_conclusion}

\appendices
\footnotesize
\input{appendices/a_architecture-ledger}
\input{appendices/b_mapping-ledger}
\input{appendices/c_recoding-packets}
\input{appendices/d_experiment-protocols}
\input{appendices/e_implementation-boundaries}
\input{appendices/f_full-evidence-ledger}

{\footnotesize
\bibliographystyle{ieeetr}
\bibliography{ref}
}

\end{document}

%% file: chapters/01_introduction.tex
\section{Introduction}

Language agents are increasingly defined by the runtime that surrounds the model: persistent state, context assembly, tool execution, lifecycle control, memory, scheduling, and recovery. Feature-level descriptions obscure this architecture. A ``memory'' component may be a checkpoint, event log, searchable store, active context, or executable skill library; a ``planner'' may be a text generator, graph traversal policy, task delegator, or commitment controller. These implementations expose different authoritative states, transition triggers, failure semantics, learning effects, and resource controls even when they use the same feature label. A useful comparison must therefore ask which state is consumed, what triggers a control action, which transition follows, what persists, and whether the outcome changes future control.

Existing reviews illuminate adjacent parts of this problem. The forty-year review catalogs cognitive architectures and their capabilities but predates modern language-agent runtimes \cite{kotseruba2018review}. CoALA imports cognitive-science concepts into a language-agent organization, while broad LLM-agent surveys organize capabilities and applications \cite{sumers2023coala,xi2023rise}. The Agent Harness Survey establishes runtime infrastructure as a research object, and a recent agentic-software review connects BDI and deliberative models to typed tools and governed execution \cite{meng2026agentharness,alenezi2026evolution}. These views establish the two shores of the comparison. They do not provide a mechanism-by-mechanism account of historical provenance, modern control-edge correspondence, implementation depth, and the residual couplings left after current specialized systems are considered. Table~\ref{tab:adjacent-review-positioning} uses five field-spanning reviews as scope comparators; specialized memory, planning, and autonomous-agent reviews informed candidate discovery but were not treated as table comparators because they do not jointly span historical architectures, runtime interfaces, and migration coding.

{\small
\begin{table}[t]
\centering
\caption{Positioning against five field-spanning adjacent reviews. ``Primary'' marks a central coding object, ``Partial'' a supporting topic, and ``Absent'' no corresponding analysis. These labels describe analytical focus rather than coverage completeness.}
\label{tab:adjacent-review-positioning}
\begin{tabular}{@{}p{0.22\textwidth}p{0.13\textwidth}p{0.13\textwidth}p{0.14\textwidth}p{0.13\textwidth}p{0.14\textwidth}@{}}
\toprule
Review & Historical mechanisms & Modern runtimes & Control-edge mapping & E/D coding & Residual bundles \\
\midrule
Forty-year cognitive-architecture review \cite{kotseruba2018review} & Primary & Absent & Partial & Absent & Absent \\
CoALA \cite{sumers2023coala} & Partial & Partial & Partial & Absent & Absent \\
General LLM-agent survey \cite{xi2023rise} & Partial & Partial & Absent & Absent & Absent \\
Agent Harness Survey \cite{meng2026agentharness} & Absent & Primary & Partial & Absent & Partial \\
Agentic software architecture review \cite{alenezi2026evolution} & Partial & Partial & Partial & Absent & Absent \\
\midrule
This review & Primary & Primary & Primary & Primary & Primary \\
\bottomrule
\end{tabular}
\end{table}
}

This review advances one bounded synthesis claim: modern language agents have migrated much of the cognitive substrate, while adaptive control mechanisms exhibit uneven migration depth. Specialized systems already implement substantial parts of memory-policy learning, typed failure recovery, team and topology adaptation, BDI control, workflow search, symbolic--learned skill coordination, resource scheduling, and uncertainty-conditioned action. Historical architectures remain useful because they specify couplings that these implementations often leave fragmented. The open research target is therefore a set of runtime invariants connecting state, failure, commitment, learning, uncertainty, and resources, rather than a wholesale reconstruction of an older architecture.

The paper makes three contributions. First, it builds a distinctive-mechanism catalog for ten cognitive architectures. Every entry identifies the consumed state, trigger, control action, state transition, learning consequence, and problem solved. Second, it introduces two independent coding axes. E1--E4 records whether a correspondence is documented lineage, structural migration, functional approximation, or convergence; D0--D4 records how deeply the specific mechanism is implemented, from a conceptual label to an experience-adapted control law. A D4 result can therefore coexist with missing historical invariants. Third, explicit residual-extraction and merge/split rules produce six candidate bundles, while adversarial closest-baseline screening closes the skill-governance candidate with GraSP and converts the five survivors into interventions with insertion points, baselines, observables, and failure conditions.

The analysis is bounded rather than exhaustive and makes no default performance claim for historical mechanisms. Sections 2--4 define the protocol and reconstruct the historical and modern strata. Section 5 reports migration depth and breakpoints. Section 6 converts the residual bundles into a runtime research agenda. Section 7 discusses validity limits and conclusions.

%% file: chapters/02_review-method.tex
\section{Review Protocol and Evidence Model}

\subsection{Review object and frozen analytical corpus}

This paper conducts a bounded mechanism-level mapping review. Its unit of analysis is a documented mechanism: an operation that consumes explicit state, is activated by an identifiable trigger, exercises control over a transition, and may change later behavior. The review does not estimate pooled effects, rank architectures on a common benchmark, or treat publication frequency as evidence of mechanism quality. Mapping is appropriate because historical architectures and language-agent systems use different terminology, implementation substrates, task domains, and evaluation protocols \cite{kotseruba2018review,sumers2023coala}.

The analytical corpus was frozen on 26 July 2026 and is divided into three strata. Corpus H contains ten historical architecture families: ACT-R, Soar, CLARION, LIDA, Hearsay-II/blackboard systems, BDI, MIDCA, ICARUS, EPIC, and Sigma. Primary theory papers, manuals, and architecture documentation establish the source mechanism and its intended operation \cite{anderson2004integrated,actrmanual,laird1987soar,soarmanual,sun2006clarion,franklin2007lida,erman1980hearsay,rao1991bdi,kinny1991commitment,rao1995bdi,cox2016midca,icarus2009,kieras1997epic,sigma2011}.

Corpus M has two modern layers. Layer M-A contains eight general runtime families used to locate implementable boundaries: Letta, LangGraph, AutoGen, AgentScope, OpenHands, Microsoft Agent Framework, the OpenAI Agents SDK, and AIOS \cite{packer2023memgpt,lettaRepo2026,langgraphRepo2026,wu2023autogen,autogenRepo2026,gao2025agentscope,wang2024openhands,wang2025the,microsoftAgentFrameworkRepo2026,openaiAgentsSdkRepo2026,mei2024aios}. Magentic-One and Agent Spec serve as supporting orchestration and specification cases \cite{fourney2024magentic,amini2025open}. Layer M-B contains forty-two mechanism-focused records used to determine the highest migration depth among reviewed cases: AgeMem, Memory-R1, MemCon, DeltaMem, DAM, A-MAC, and CURATOR for memory; PALADIN, AgentHER, Reflexion, AgentDebugX, and Shepherd for failure and execution state; DyLAN, adaptive graph pruning, and Global Workspace Agents for competitive selection and broadcast; hybrid and self-aware BDI--LLM systems, Devil's Advocate, Cognitive Control Architecture, goal drift, and premature commitment; ADAS, AFlow, and MetaReflection for meta-level adaptation; SGDR, SCALAR, Voyager, Agent Workflow Memory, SkillComposer, SkillOps, GraSP, Agentic Compilation, SkVM, SkillSmith, and SkCC for skill induction, composition, governance, and compilation; AIOS, AgentRM, and budget-aware value search for resources; and UALA, KnowNo, Calibrate-Then-Act, and Utility-Guided Agent Orchestration for uncertainty-conditioned control \cite{yu2026agemem,yan2025memoryr1,jiang2026memcon,zhang2026deltamem,sun2025dam,zhang2026amac,wu2026curator,vuddanti2025paladin,ding2026agenther,shinn2023reflexion,zhu2026agentdebugx,yu2026shepherd,liu2023dylan,li2025agp,shang2026gwa,owayyed2025controlled,lawton2026selfaware,wang2024devilsadvocate,liang2025cca,arike2025goaldrift,mehta2026premature,hu2024adas,zhang2024aflow,gupta2024metareflection,li2026sgdr,zabounidis2026scalar,wang2023voyager,wang2024awm,zhao2026skillcomposer,song2026skillops,xia2026grasp,chundru2026agenticcompilation,chen2026skvm,xu2026skillsmith,ouyang2026skcc,mei2024aios,she2026agentrm,li2026bavt,han2024uala,ren2023knowno,ding2026cta,liu2026utility}. AIOS is intentionally cross-indexed because it is both a general runtime proposal and a resource-governance implementation; counts describe analytical roles rather than disjoint publications.

\subsection{Search and screening procedure}

Candidate generation combined arXiv title/abstract search, Semantic Scholar exact-title and citation chasing, primary-paper reference snowballing, and official repository inspection for runtime interfaces. Six query blocks mirrored the residual mechanism families: adaptive memory policy; failure diagnosis, reversible branching, recovery, and compilation; global workspace, broadcast, and competition; BDI intention, intent graph, commitment, and method switching; skill induction, compilation, applicability, and fallback; and uncertainty-conditioned resource allocation, interruption, and stopping. Searches used the conjunction ``LLM agent'' with each block's terms, admitted records dated no later than the freeze date, and retained a case only when full text exposed a state object plus a control, transition, or learning edge. A dated query-and-disposition log supplies replayable query templates and record-level dispositions, while Appendix~\ref{app:full-evidence-ledger} reports the included counterexamples and their effect on each residual. The procedure is documented and replayable at the query-template level. Result counts were not frozen, so no search-recall or stable hit-count claim is made.

Corpus B contains explicit historical--modern bridges. CoALA, LLM-ACTR, Bootstrapping Cognitive Agents, and the hybrid BDI--LLM system support direct lineage or integration claims \cite{sumers2023coala,cognitivellm2024,bootstrapping2024,owayyed2025controlled}. Similarity inferred only by comparing Corpus H and Corpus M is coded as approximation or convergence. Records are included when they expose implementable state, control, transition, learning, failure, uncertainty, or resource semantics. Generic capability claims, unversioned product descriptions, standalone retrieval augmentation, performance results without mechanism exposure, and brain--Transformer analogies without an agent operation are excluded from mechanism coding.

\subsection{Seven-field mechanism representation}

Each mechanism is represented by
\[
\mathcal{K}=\langle S,C,T,P,F,L,R\rangle ,
\]
where $S$ is the operative state substrate, $C$ the control locus, $T$ the transition trigger, $P$ the persistence boundary, $F$ the failure semantics, $L$ the learning operator, and $R$ the resource or governance policy. The tuple forces mechanisms with a shared purpose to reveal differences in authority and dynamics. A searchable store and ACT-R declarative memory may both supply past information, for example, while differing in activation, latency, decay, retrieval competition, and utility coupling.

For every correspondence, historical and modern instances are coded side by side. The evidence packet records: (1) the historical mechanism and primary source; (2) the modern implementation and primary source; (3) matching state object; (4) matching trigger; (5) matching control or learning edge; (6) preserved invariant; (7) missing invariant; and (8) official implementation evidence when the mechanism depends on executable behavior. Repository popularity is never used as scientific evidence. A repository is admissible only as implementation evidence when a release, tag, or commit identifies the analyzed state.

\subsection{Independent evidence and depth axes}

Evidence relation and migration depth answer different questions and are coded independently. E1 identifies an explicit source-cited lineage. E2 identifies an implemented structural correspondence. E3 identifies a functionally comparable control solution with materially different internal organization. E4 identifies structural convergence without evidence of transfer. Migration depth applies to the specific mapped mechanism, never to an entire architecture or framework. D0 marks conceptual resemblance; D1 a similar callable interface; D2 explicit state and component structure; D3 a trigger that autonomously executes the mapped control transition; and D4 an outcome-dependent update to future control policy.

Each correspondence receives exactly one E-code. E-codes are relation types rather than maturity levels, so interval notation is prohibited for a single correspondence. When several descriptions appear applicable, coding follows an evidentiary precedence rule: E1 when the modern source explicitly attributes the mapped mechanism to the historical architecture; otherwise E2 when implemented source-like state and control structure are present; otherwise E3 when the same control problem is solved through materially different organization; otherwise E4 when node--edge similarity is inferred without transfer evidence. Rows that aggregate heterogeneous modern cases must list case-specific codes or be split.

{\scriptsize
\begin{table}[t]
\centering
\caption{Operational coding rules for evidence relation and migration depth.}
\label{tab:evidence-depth-rubric}
\begin{tabular}{@{}p{0.05\textwidth}p{0.17\textwidth}p{0.27\textwidth}p{0.26\textwidth}p{0.11\textwidth}@{}}
\toprule
Code & Question & Necessary condition & Exclusion test & Example \\
\midrule
E1 & Documented lineage & Modern source names the historical architecture and connects it to the mapped mechanism. & A bibliography mention without a mechanism claim is insufficient. & CoALA \\
E2 & Structural migration & Implemented state plus a corresponding control, transition, or learning edge. & Comparable outputs alone are insufficient. & Hybrid BDI--LLM \\
E3 & Functional approximation & Comparable control problem and observable function with different internal organization. & Shared terminology alone is insufficient. & PALADIN vs. Soar \\
E4 & Structural convergence & Node-and-edge organization is comparable, with no transfer evidence found. & Lack of lineage evidence must be stated explicitly. & AIOS vs. EPIC \\
\midrule
D0 & Conceptual resemblance & A comparable idea or label can be identified. & No callable interface or persistent mechanism state. & ``Reflection'' label \\
D1 & Interface correspondence & The runtime exposes a callable operation with a comparable role. & No explicit mapped state-transition organization. & Memory API \\
D2 & State/component structure & Mapped state objects and components are explicit. & Trigger does not automatically exercise the historical control edge. & Stored intention record \\
D3 & Runtime control law & An explicit trigger executes the mapped control action and changes state. & Outcome does not update later selection policy. & KnowNo help trigger \\
D4 & Adaptive control law & Experience updates the future invocation, selection, or control policy. & Logging or storing text without policy change is insufficient. & AgeMem; SCALAR \\
\bottomrule
\end{tabular}
\end{table}
}

A high D-level does not imply complete historical migration. SCALAR can reach D4 for bidirectional symbolic--policy updating while still lacking cross-domain portability, calibrated invocation, and a reliable explicit-planning fallback. Conversely, an E1 lineage claim may remain D1 if it cites an architecture but implements only a similar interface. The paper therefore reports each result as an ordered claim packet: historical mechanism $\rightarrow$ modern implementation $\rightarrow$ state $\rightarrow$ trigger $\rightarrow$ control edge $\rightarrow$ preserved invariant $\rightarrow$ missing invariant $\rightarrow$ E-level $\rightarrow$ D-level.

Residual bundles are generated from the coded breakpoints rather than chosen directly from architecture names. First, every mapping row contributes its smallest missing state, trigger, authority, transition, or learning invariant. Atomic residuals are merged only when they share a runtime insertion boundary and one atom supplies state or authority required by the other. They are split when they have an independent trigger, baseline, or falsifier. A bundle is retained only when every constituent edge has a reviewed modern precedent, the coupling is absent from the frozen corpus, and the composition admits a concrete insertion point and rejection condition. Section~\ref{sec:residual-bundles} records the resulting partition and the nearest rejected alternatives. The rules initially produced six candidates. Closest-baseline screening then found GraSP implementing the complete B5 selection--verification--repair--fallback chain, so B5 is retained as a closed negative case and five candidates remain residual. This partition is auditable conditional on the current provisional codes, not a claim that five is the unique partition of the literature.

\subsection{Claim adjudication and validity boundary}

A mapping passes the claim test only when it contains node correspondence, edge correspondence, and at least one preserved invariant. Shared nouns, visually similar diagrams, or comparable outputs fail when the authoritative state or control path cannot be identified. Strong D3 and D4 claims additionally require a historical primary source, a modern primary paper, implementation evidence when available, and an explicit missing invariant. The reported score is the highest-depth instance admitted by the recorded search and screening procedure, not a prevalence estimate or a universal maximum.

Coding proceeded in two passes: extraction of the seven fields followed by adversarial review of the claimed edge, preserved invariant, and missing invariant. The present manuscript does not claim independent inter-rater reliability. Its evidence packets and boundary cases are retained for a second-author recoding of all D3/D4 claims before submission; disagreements must be reported and adjudicated without raising a score when implementation evidence is ambiguous. This disclosure separates a reproducible protocol from reliability evidence that has not yet been produced.

The review makes no PRISMA completeness claim and no causal performance claim. A residual gap is admitted only after current D3/D4 precedents are accounted for and the remaining property can be isolated against the closest baseline. Historical value is therefore expressed as a falsifiable control hypothesis, not as an assumption that an older mechanism will improve reliability, efficiency, generality, or autonomy.

\input{figures/review-pipeline-clean}

%% file: figures/review-pipeline-clean.tex
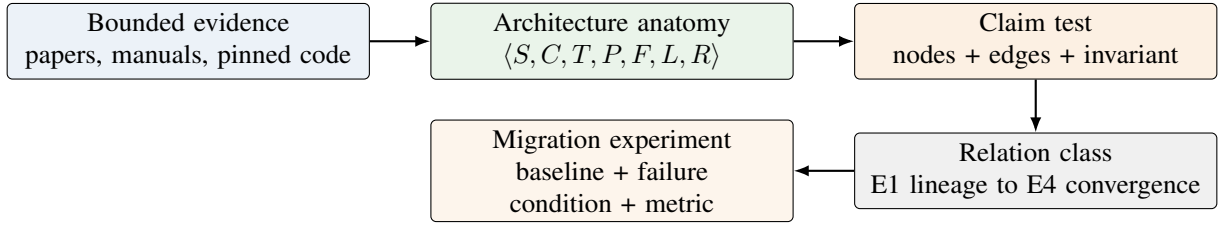
\begin{figure}[t]
\centering
\begin{tikzpicture}[
  stage/.style={draw, rounded corners=2pt, align=center, text width=0.25\textwidth, minimum height=10mm, font=\small},
  flow/.style={-{Latex[length=2mm]}, thick},
  node distance=7mm and 8mm
]
\node[stage, fill=accentblue!10] (evidence) {Bounded evidence\\papers, manuals, pinned code};
\node[stage, fill=accentgreen!10, right=of evidence] (anatomy) {Architecture anatomy\\$\langle S,C,T,P,F,L,R\rangle$};
\node[stage, fill=accentorange!12, right=of anatomy] (claim) {Claim test\\nodes + edges + invariant};
\node[stage, fill=gray!12, below=of claim] (relation) {Relation class\\E1 lineage to E4 convergence};
\node[stage, fill=accentorange!8, left=of relation] (experiment) {Migration experiment\\baseline + failure condition + metric};
\draw[flow] (evidence) -- (anatomy);
\draw[flow] (anatomy) -- (claim);
\draw[flow] (claim) -- (relation);
\draw[flow] (relation) -- (experiment);
\end{tikzpicture}
\caption{Review logic. A similarity claim is admitted only after architecture reconstruction and a node--edge--invariant test; migration gaps become experiments rather than inheritance claims.}
\label{fig:review-pipeline}
\end{figure}

%% file: chapters/03_historical-architectures.tex
\section{Cognitive Mechanism Catalog}
\label{sec:mechanism-catalog}

The historical corpus is used as a catalog of executable control mechanisms. Architectural names alone are too coarse: ACT-R and Soar are both production-based, yet one couples typed buffers to activation and utility while the other couples operator selection to impasses, substates, and chunking. We therefore summarize each architecture through a fixed operational chain:
\[
\text{state consumed}\rightarrow\text{trigger}\rightarrow\text{control action}
\rightarrow\text{state transition}\rightarrow\text{learning consequence}.
\]
The final question is practical: which recurrent control problem does the chain solve? Table~\ref{tab:signature-mechanisms} gives the compact catalog; Appendix~\ref{app:architecture-ledger} records the complete $S/C/T/P/F/L/R$ coding.

{\scriptsize
\begin{longtable}{@{}p{0.09\textwidth}p{0.19\textwidth}p{0.28\textwidth}p{0.19\textwidth}p{0.19\textwidth}@{}}
\caption{Distinctive mechanisms, their operation, and the control problem they solve.}\label{tab:signature-mechanisms}\\
\toprule
Architecture & Signature mechanism & Runtime operation & Learning consequence & Problem solved \\
\midrule
\endfirsthead
\toprule
Architecture & Signature mechanism & Runtime operation & Learning consequence & Problem solved \\
\midrule
\endhead
\bottomrule
\endfoot
ACT-R & Typed buffers; activation; production utility & Buffer state triggers production competition; retrieval latency depends on activation; utility selects an applicable action. & Accessibility and utility change future retrieval and rule choice. & Bounded access to memory and value-sensitive action under timing constraints. \\
\midrule
Soar & Operators; typed impasses; substates; chunking & Unresolved operator choice creates a scoped child problem; its result returns to the parent. & The resolution is compiled into a production. & Recover from a specific control failure without repeating the same search. \\
\midrule
CLARION & Explicit--implicit levels; cross-level learning & Rules and learned action tendencies jointly propose behavior; either level may guide the other. & Rules can be extracted from skills and skills trained from rules. & Combine flexible deliberation with low-cost practiced behavior. \\
\midrule
LIDA & Codelet competition; limited workspace; broadcast & Coalitions compete for scarce workspace access; the winner is broadcast to memory and action processes. & Broadcast outcomes reinforce later admission and action. & Coordinate specialized processes under attention limits. \\
\midrule
Hearsay-II & Blackboard; knowledge sources; agenda & A posted partial hypothesis enables knowledge sources; a scheduler chooses the next contribution. & Classic Hearsay-II primarily accumulates structured partial solutions. & Opportunistically combine heterogeneous expertise without a fixed pipeline. \\
\midrule
BDI & Options; intention commitment; reconsideration & Beliefs and desires generate options; filtering creates intentions that persist until a reconsideration condition. & Learning is optional, while commitment changes future deliberation. & Prevent goal drift and endless replanning during long-horizon action. \\
\midrule
MIDCA & Object/meta cycles; trace diagnosis; method control & A meta-cycle inspects object-level traces, diagnoses failure, and changes the active reasoning method. & Successful method changes can inform later meta-control. & Give monitoring causal authority over planning and execution. \\
\midrule
ICARUS & Grounded concepts; hierarchical skills & Percepts instantiate concepts; goals select skills whose subskills or actions advance the state. & Skill structures can be acquired or refined from problem solving. & Turn symbolic goals into reusable, environment-grounded procedures. \\
\midrule
EPIC & Parallel processors; timing and bottlenecks & Perceptual, cognitive, and motor processors run with explicit temporal constraints and shared bottlenecks. & Learning is secondary to architectural timing commitments. & Predict interference, latency, and feasible parallelism. \\
\midrule
Sigma & Factor graph; message passing; decision control & Probabilistic and symbolic constraints exchange messages; decisions use the resulting belief state. & Shared graphical representations support parameter and structure updates. & Propagate uncertainty through perception, inference, and action. \\
\end{longtable}
}

\subsection{Selection, retrieval, and failure: ACT-R and Soar}

\paragraph{ACT-R.}
ACT-R exposes a small typed working state through goal, retrieval, perceptual, and motor buffers. A buffer match triggers production competition; the selected production changes a buffer, requests a module operation, or initiates action. Declarative chunks persist beyond the current cycle, but their availability and retrieval latency depend on activation. Learned production utility separates logical applicability from expected desirability \cite{anderson2004integrated,actrmanual}. The signature mechanism therefore couples three decisions often separated in software agents: what enters working state, how long retrieval takes, and which action is worth executing. It solves bounded memory access and value-sensitive control rather than generic storage.

\paragraph{Soar.}
Soar represents the current problem in working memory and proposes operators as candidate state transitions. Preferences select among operators. A tie, conflict, rejection, or missing choice is represented as a typed impasse, which creates a child substate with its own local decision process. The result is returned to the parent, and chunking can compile the dependency conditions and resolution into a future production \cite{laird1987soar,soarmanual,soarchunking}. The distinctive chain is thus failure type $\rightarrow$ scoped recovery $\rightarrow$ reusable control knowledge. Retry or reflection captures only a fragment when it lacks the typed parent--child state and compilation edge.

\input{figures/historical-actr-soar-clean}

\subsection{Representation and workspace coordination}

\paragraph{CLARION.}
CLARION maintains explicit rules and implicit learned tendencies as different knowledge forms. Current context activates candidates at both levels; cross-level integration chooses behavior. Bottom-up extraction makes a successful implicit regularity explicit, while top-down assimilation trains lower-level behavior from instruction or rules \cite{sun2006clarion}. This arrangement solves the cost--flexibility tradeoff between deliberate reasoning and practiced execution. A stored workflow resembles only the explicit side until the runtime can invoke it cheaply, estimate applicability, and fall back when the skill is unsafe or out of distribution.

\paragraph{LIDA.}
LIDA organizes many specialized codelets into coalitions that compete for access to a capacity-limited global workspace. A winner is broadcast to memory, action selection, and learning processes; the broadcast changes which processes can act in the next cycle \cite{franklin2007lida}. The mechanism solves content-level coordination under attention limits. A message bus provides transport, while a LIDA-like workspace additionally requires scarcity, competition, a winner, broad availability of the selected content, and learning from admission outcomes.

\paragraph{Hearsay-II and blackboard control.}
Hearsay-II stores layered partial hypotheses on a shared blackboard. A new hypothesis or state change enables knowledge sources, and a scheduler selects which enabled source receives control next \cite{erman1980hearsay}. The mechanism supports opportunistic problem solving when no fixed pipeline can predict the useful order of expertise. Its distinctive object is the evolving partial solution, not the message channel. Modern shared state structurally converges only when contributions modify a common problem representation and scheduling depends on that representation.

\input{figures/historical-hybrid-clean}

\subsection{Commitment, meta-control, and grounded skills}

\paragraph{BDI.}
BDI agents update beliefs from observations, generate options from desires, filter options into intentions, and execute plans associated with those intentions. Intentions constrain later deliberation until a commitment strategy releases them; blind, single-minded, open-minded, bold, and cautious variants differ in how success, impossibility, goal relevance, and the cost of reconsideration affect release \cite{rao1991bdi,kinny1991commitment,rao1995bdi}. The mechanism solves goal drift and deliberation thrashing. A handoff or task field lacks this semantics unless the runtime records why the commitment remains active and which event authorizes continuation, delegation, suspension, or abandonment.

\paragraph{MIDCA.}
MIDCA separates an object-level cognitive cycle from a meta-level cycle that reads traces of the object cycle. A detected anomaly triggers diagnosis; the meta-controller may change the planner, strategy, goal handling, or another object-level method \cite{cox2016midca}. Monitoring becomes consequential because it has method authority. A log, evaluator, or guardrail is therefore only a substrate until its diagnosis can select a different reasoning process during execution.

\paragraph{ICARUS.}
ICARUS grounds symbolic concepts in perceptual state and represents skills as hierarchical structures linking goals to subgoals and primitive actions. A goal and recognized situation trigger an applicable skill; execution descends through its subskills until an environment action changes the state \cite{icarus2009}. The mechanism solves grounded procedural reuse. Its migration test is stronger than storing text: the skill needs explicit applicability conditions, effects, executable decomposition, and revision from failure.

\subsection{Resources and uncertainty: EPIC and Sigma}

\paragraph{EPIC.}
EPIC models perceptual, cognitive, and motor processors with explicit processing times, parallel pathways, and bottlenecks \cite{kieras1997epic}. A perceptual event initiates processor activity; shared bottlenecks and motor timing constrain which operations overlap and which must wait. Its contribution is a resource law that predicts latency and interference, rather than a generic list of modules. The modern analogue must therefore make scheduling, preemption, and parallelism depend on task state and resource demand.

\paragraph{Sigma.}
Sigma uses factor graphs as a shared substrate for symbolic, probabilistic, and decision computations. Evidence triggers message passing; updated beliefs support action selection, and learning can update parameters or structures in the same representation \cite{sigma2011}. The mechanism solves cross-stage uncertainty propagation. A confidence score at one tool boundary is a local control signal; Sigma's stronger invariant is that uncertainty remains part of state as it moves through inference and decision.

\input{figures/historical-resource-uncertainty-clean}

Across the catalog, the reusable asset is the coupling among mechanisms. Buffers matter because activation and utility govern their use; impasses matter because they create substates and learning; workspaces matter because capacity, competition, broadcast, and learning form one cycle. These couplings become the unit of migration analysis in Section~\ref{sec:migration-depth}.

%% file: figures/historical-actr-soar-clean.tex
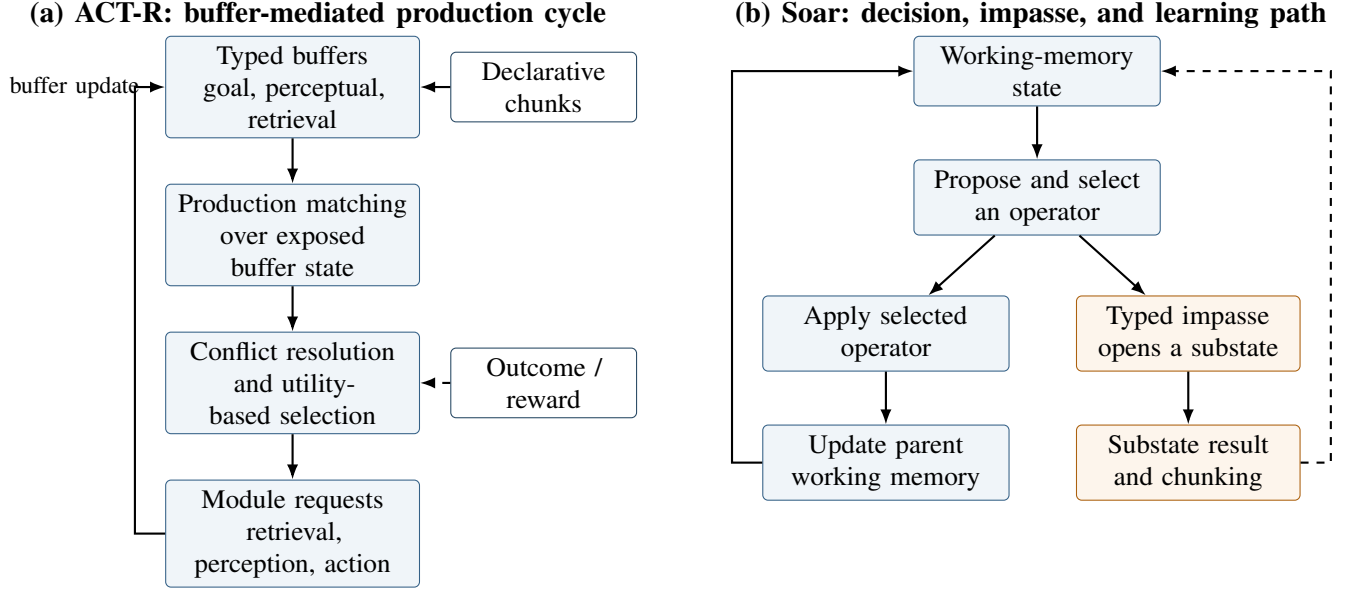
\begin{figure*}[t]
\centering
\begin{minipage}[t]{0.48\textwidth}
\centering
\textbf{(a) ACT-R: buffer-mediated production cycle}\par\smallskip
\begin{tikzpicture}[
  hist/.style={draw=accentblue!75!black, fill=accentblue!8, rounded corners=2pt, align=center, text width=31mm, minimum height=9mm, font=\small},
  side/.style={draw=accentblue!55!black, fill=white, rounded corners=2pt, align=center, text width=22mm, minimum height=8mm, font=\small},
  flow/.style={-{Latex[length=2mm]}, thick},
  learn/.style={-{Latex[length=2mm]}, thick, dashed},
  node distance=6mm and 7mm
]
\node[hist] (buffers) {Typed buffers\\goal, perceptual, retrieval};
\node[hist, below=of buffers] (match) {Production matching\\over exposed buffer state};
\node[hist, below=of match] (select) {Conflict resolution\\and utility-based selection};
\node[hist, below=of select] (modules) {Module requests\\retrieval, perception, action};
\node[side, right=4mm of buffers] (dm) {Declarative\\chunks};
\node[side, right=4mm of select] (reward) {Outcome /\\reward};
\draw[flow] (buffers) -- (match);
\draw[flow] (match) -- (select);
\draw[flow] (select) -- (modules);
\draw[flow] (dm) -- (buffers);
\draw[flow] (modules.west) -- ++(-4mm,0) |- node[pos=0.75, left, font=\footnotesize]{buffer update} (buffers.west);
\draw[learn] (reward) -- (select);
\end{tikzpicture}
\end{minipage}\hfill
\begin{minipage}[t]{0.48\textwidth}
\centering
\textbf{(b) Soar: decision, impasse, and learning path}\par\smallskip
\begin{tikzpicture}[
  hist/.style={draw=accentblue!75!black, fill=accentblue!8, rounded corners=2pt, align=center, text width=30mm, minimum height=9mm, font=\small},
  branch/.style={draw=accentorange!75!black, fill=accentorange!8, rounded corners=2pt, align=center, text width=27mm, minimum height=9mm, font=\small},
  flow/.style={-{Latex[length=2mm]}, thick},
  learn/.style={-{Latex[length=2mm]}, thick, dashed},
  node distance=7mm and 8mm
]
\node[hist] (wm) {Working-memory\\state};
\node[hist, below=of wm] (decision) {Propose and select\\an operator};
\node[hist] (apply) at ([xshift=-20mm,yshift=-18mm]decision) {Apply selected\\operator};
\node[branch] (substate) at ([xshift=20mm,yshift=-18mm]decision) {Typed impasse\\opens a substate};
\node[hist, below=of apply] (update) {Update parent\\working memory};
\node[branch, below=of substate] (resolve) {Substate result\\and chunking};
\draw[flow] (wm) -- (decision);
\draw[flow] (decision) -- (apply);
\draw[flow] (decision) -- (substate);
\draw[flow] (apply) -- (update);
\draw[flow] (substate) -- (resolve);
\draw[flow] (update.west) -- ++(-4mm,0) |- (wm.west);
\draw[learn] (resolve.east) -- ++(4mm,0) |- (wm.east);
\end{tikzpicture}
\end{minipage}
\caption{Executable control cycles for ACT-R and Soar. ACT-R exposes typed buffers to production matching and updates retrieval and utility through module outcomes. Soar branches from operator selection into either normal application or an impasse-created substate whose resolution can be compiled by chunking \cite{actrmanual,soarmanual,soarchunking}.}
\label{fig:actr-soar-cycles}
\end{figure*}

%% file: figures/historical-hybrid-clean.tex
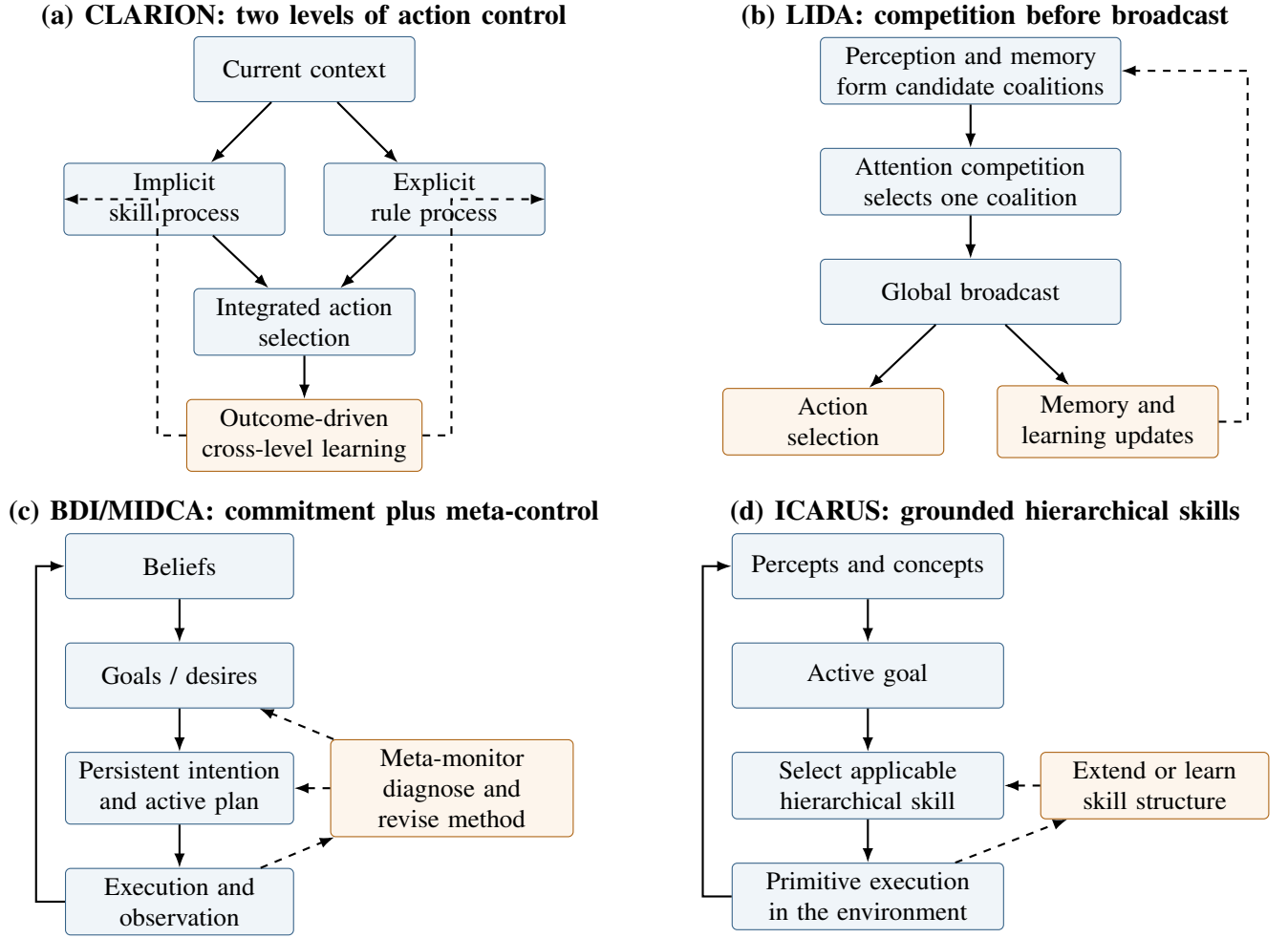
\begin{figure*}[t]
\centering
\begin{minipage}[t]{0.48\textwidth}
\centering\textbf{(a) CLARION: two levels of action control}\par\smallskip
\begin{tikzpicture}[node distance=6mm and 9mm,
  box/.style={draw=accentblue!75!black, fill=accentblue!8, rounded corners=2pt, align=center, text width=28mm, minimum height=9mm, font=\small},
  learn/.style={draw=accentorange!75!black, fill=accentorange!8, rounded corners=2pt, align=center, text width=30mm, minimum height=9mm, font=\small},
  flow/.style={-{Latex[length=2mm]}, thick}, feedback/.style={-{Latex[length=2mm]}, dashed, thick}]
\node[box] (context) {Current context};
\node[box] (implicit) at ([xshift=-18mm,yshift=-18mm]context) {Implicit\\skill process};
\node[box] (explicit) at ([xshift=18mm,yshift=-18mm]context) {Explicit\\rule process};
\node[box] (action) at ([xshift=18mm,yshift=-17mm]implicit) {Integrated action\\selection};
\node[learn, below=of action] (learning) {Outcome-driven\\cross-level learning};
\draw[flow] (context) -- (implicit);
\draw[flow] (context) -- (explicit);
\draw[flow] (implicit) -- (action);
\draw[flow] (explicit) -- (action);
\draw[flow] (action) -- (learning);
\draw[feedback] (learning.west) -- ++(-4mm,0) |- (implicit.west);
\draw[feedback] (learning.east) -- ++(4mm,0) |- (explicit.east);
\end{tikzpicture}
\end{minipage}\hfill
\begin{minipage}[t]{0.48\textwidth}
\centering\textbf{(b) LIDA: competition before broadcast}\par\smallskip
\begin{tikzpicture}[node distance=6mm,
  box/.style={draw=accentblue!75!black, fill=accentblue!8, rounded corners=2pt, align=center, text width=39mm, minimum height=9mm, font=\small},
  split/.style={draw=accentorange!75!black, fill=accentorange!8, rounded corners=2pt, align=center, text width=28mm, minimum height=9mm, font=\small},
  flow/.style={-{Latex[length=2mm]}, thick}, feedback/.style={-{Latex[length=2mm]}, dashed, thick}]
\node[box] (candidates) {Perception and memory\\form candidate coalitions};
\node[box, below=of candidates] (competition) {Attention competition\\selects one coalition};
\node[box, below=of competition] (broadcast) {Global broadcast};
\node[split] (action) at ([xshift=-19mm,yshift=-18mm]broadcast) {Action\\selection};
\node[split] (memory) at ([xshift=19mm,yshift=-18mm]broadcast) {Memory and\\learning updates};
\draw[flow] (candidates) -- (competition);
\draw[flow] (competition) -- (broadcast);
\draw[flow] (broadcast) -- (action);
\draw[flow] (broadcast) -- (memory);
\draw[feedback] (memory.east) -- ++(4mm,0) |- (candidates.east);
\end{tikzpicture}
\end{minipage}

\vspace{4mm}
\begin{minipage}[t]{0.48\textwidth}
\centering\textbf{(c) BDI/MIDCA: commitment plus meta-control}\par\smallskip
\begin{tikzpicture}[node distance=6mm and 8mm,
  box/.style={draw=accentblue!75!black, fill=accentblue!8, rounded corners=2pt, align=center, text width=29mm, minimum height=9mm, font=\small},
  meta/.style={draw=accentorange!75!black, fill=accentorange!8, rounded corners=2pt, align=center, text width=31mm, minimum height=9mm, font=\small},
  flow/.style={-{Latex[length=2mm]}, thick}, control/.style={-{Latex[length=2mm]}, dashed, thick}]
\node[box] (belief) {Beliefs};
\node[box, below=of belief] (goal) {Goals / desires};
\node[box, below=of goal] (intent) {Persistent intention\\and active plan};
\node[box, below=of intent] (action) {Execution and\\observation};
\node[meta, right=5mm of intent] (monitor) {Meta-monitor\\diagnose and revise method};
\draw[flow] (belief) -- (goal);
\draw[flow] (goal) -- (intent);
\draw[flow] (intent) -- (action);
\draw[flow] (action.west) -- ++(-4mm,0) |- (belief.west);
\draw[control] (action) -- (monitor);
\draw[control] (monitor) -- (intent);
\draw[control] (monitor) -- (goal);
\end{tikzpicture}
\end{minipage}\hfill
\begin{minipage}[t]{0.48\textwidth}
\centering\textbf{(d) ICARUS: grounded hierarchical skills}\par\smallskip
\begin{tikzpicture}[node distance=6mm and 8mm,
  box/.style={draw=accentblue!75!black, fill=accentblue!8, rounded corners=2pt, align=center, text width=35mm, minimum height=9mm, font=\small},
  learn/.style={draw=accentorange!75!black, fill=accentorange!8, rounded corners=2pt, align=center, text width=29mm, minimum height=9mm, font=\small},
  flow/.style={-{Latex[length=2mm]}, thick}, feedback/.style={-{Latex[length=2mm]}, dashed, thick}]
\node[box] (percepts) {Percepts and concepts};
\node[box, below=of percepts] (goals) {Active goal};
\node[box, below=of goals] (skills) {Select applicable\\hierarchical skill};
\node[box, below=of skills] (execute) {Primitive execution\\in the environment};
\node[learn, right=5mm of skills] (learning) {Extend or learn\\skill structure};
\draw[flow] (percepts) -- (goals);
\draw[flow] (goals) -- (skills);
\draw[flow] (skills) -- (execute);
\draw[flow] (execute.west) -- ++(-4mm,0) |- (percepts.west);
\draw[feedback] (execute) -- (learning);
\draw[feedback] (learning) -- (skills);
\end{tikzpicture}
\end{minipage}
\caption{Four distinct execution organizations. CLARION coordinates explicit and implicit control; LIDA requires competition before selective broadcast; BDI/MIDCA separates persistent task commitment from meta-level diagnosis; ICARUS selects hierarchical skills whose applicability is grounded in percepts and environmental action \cite{sun2006clarion,franklin2007lida,cox2016midca,icarus2009}.}
\label{fig:hybrid-architecture-cycles}
\end{figure*}

%% file: figures/historical-resource-uncertainty-clean.tex
\begin{figure*}[t]
\centering
\begin{minipage}[t]{0.48\textwidth}
\centering\textbf{(a) EPIC: concurrent processors with bottlenecks}\par\smallskip
\begin{tikzpicture}[node distance=6mm and 8mm,
  box/.style={draw=accentblue!75!black, fill=accentblue!8, rounded corners=2pt, align=center, text width=31mm, minimum height=9mm, font=\small},
  limit/.style={draw=accentorange!75!black, fill=accentorange!8, rounded corners=2pt, align=center, text width=30mm, minimum height=9mm, font=\small},
  flow/.style={-{Latex[length=2mm]}, thick}, timing/.style={-{Latex[length=2mm]}, dashed, thick}]
\node[box] (percept) {Visual and auditory\\processors};
\node[box, below=of percept] (stores) {Perceptual stores\\and task state};
\node[limit, below=of stores] (cognitive) {Production-rule\\cognitive processor};
\node[box, below=of cognitive] (motor) {Manual and vocal\\motor processors};
\node[limit, right=5mm of cognitive] (clock) {Durations, queues,\\and channel capacity};
\draw[flow] (percept) -- (stores);
\draw[flow] (stores) -- (cognitive);
\draw[flow] (cognitive) -- (motor);
\draw[timing] (clock) -- (percept);
\draw[timing] (clock) -- (cognitive);
\draw[timing] (clock) -- (motor);
\end{tikzpicture}
\end{minipage}\hfill
\begin{minipage}[t]{0.48\textwidth}
\centering\textbf{(b) Sigma: inference and action in a factor graph}\par\smallskip
\begin{tikzpicture}[node distance=6mm and 8mm,
  box/.style={draw=accentblue!75!black, fill=accentblue!8, rounded corners=2pt, align=center, text width=32mm, minimum height=9mm, font=\small},
  factor/.style={draw=accentorange!75!black, fill=accentorange!8, rounded corners=2pt, align=center, text width=31mm, minimum height=9mm, font=\small},
  flow/.style={-{Latex[length=2mm]}, thick}, learn/.style={-{Latex[length=2mm]}, dashed, thick}]
\node[box] (evidence) {Evidence and\\current variables};
\node[factor, below=of evidence] (graph) {Factors encode rules,\\memory, and constraints};
\node[box, below=of graph] (inference) {Message passing updates\\belief distributions};
\node[factor, below=of inference] (decision) {Expected utility\\and action selection};
\node[box, right=5mm of inference] (outcome) {Action outcome\\and learning signal};
\draw[flow] (evidence) -- (graph);
\draw[flow] (graph) -- (inference);
\draw[flow] (inference) -- (decision);
\draw[flow] (decision) -- (outcome);
\draw[learn] (outcome) -- (graph);
\end{tikzpicture}
\end{minipage}
\caption{Two additional control organizations. EPIC makes processor overlap, duration, and contention explicit; Sigma propagates evidence through a factor graph so uncertainty and utility can influence action and learning \cite{kieras1997epic,sigma2011}.}
\label{fig:epic-sigma-cycles}
\end{figure*}

%% file: chapters/04_modern-runtimes.tex
\section{Modern Agent Runtime Landscape}
\label{sec:modern-landscape}

The modern corpus is separated into general runtimes and mechanism-focused systems. General runtimes reveal where a control mechanism could be inserted; specialized systems establish how deeply a particular mechanism has already been implemented. Mixing these layers would produce two errors: a framework API could be mistaken for a learned control law, while a research prototype could be mistaken for a portable production invariant.

\subsection{Layer A: general runtime boundaries}

Letta makes memory management and long-lived agent state first-class runtime concerns. LangGraph represents execution as a stateful graph with checkpoint, interrupt, and resume boundaries. AutoGen Core and AgentScope provide event, message, agent, model, and tool interfaces for orchestrated execution. OpenHands uses an event-oriented agent--computer interaction loop and exposes executable action boundaries. Microsoft Agent Framework and the OpenAI Agents SDK expose workflow, run-state, handoff, approval, guardrail, tool, cancellation, and resumption boundaries. AIOS proposes an operating-system layer with scheduling, context, memory, storage, tool, and access-control managers \cite{packer2023memgpt,lettaRepo2026,langgraphRepo2026,wu2023autogen,autogenRepo2026,gao2025agentscope,wang2024openhands,wang2025the,microsoftAgentFrameworkRepo2026,openaiAgentsSdkRepo2026,mei2024aios}.

\input{figures/modern-runtime-clean}

These systems provide genuine engineering state and control surfaces. Their released interfaces support persistence, replay, tool governance, interruption, or orchestration. The interfaces do not by themselves determine which memory should be admitted, why a failure creates a specific subproblem, when an intention should survive a new model output, or how uncertainty should reallocate budget. The distinction is authority: a checkpoint stores state, while a cognitive control mechanism defines a trigger and transition law over that state.

\subsection{Layer B: specialized control implementations}

The mechanism-focused layer changes the gap analysis. AgeMem, Memory-R1, MemCon, DeltaMem, and A-MAC implement adaptive memory policies in which outcomes or cross-validated optimization alter later memory operations or admission \cite{yu2026agemem,yan2025memoryr1,jiang2026memcon,zhang2026deltamem,zhang2026amac}; CURATOR fits online helpfulness and retrieval propensity, then uses net value minus harm per byte to govern keep, share, and trust under physical budgets \cite{wu2026curator}; DAM supplies a decision-theoretic design framework without an evaluated control algorithm \cite{sun2025dam}. PALADIN maps failure types to recovery, AgentDebugX integrates diagnosis and rerun, Shepherd exposes reversible first-class execution traces, and AgentHER and Reflexion change later behavior from failures \cite{vuddanti2025paladin,zhu2026agentdebugx,yu2026shepherd,ding2026agenther,shinn2023reflexion}. DyLAN and adaptive graph pruning learn team or edge admission, while Global Workspace Agents implements limited working state, attention selection, broadcast, and proposal write-back without outcome-trained content admission \cite{liu2023dylan,li2025agp,shang2026gwa}. Hybrid and self-aware BDI--LLM systems preserve explicit deliberation loops; Devil's Advocate revises plans; Cognitive Control Architecture uses an intent graph, deviation trigger, and adjudicator; goal-drift and premature-commitment studies expose complementary failure modes \cite{owayyed2025controlled,lawton2026selfaware,wang2024devilsadvocate,liang2025cca,arike2025goaldrift,mehta2026premature}.

Meta-level adaptation also has concrete precedents. ADAS and AFlow search over code-represented agents or workflows using evaluation feedback, while MetaReflection learns reusable instructions from past reflections \cite{hu2024adas,zhang2024aflow,gupta2024metareflection}. Voyager compiles verified behaviors into reusable skills; SGDR and Agent Workflow Memory induce reusable workflows; SCALAR links explicit specifications to learned policies; SkillComposer learns task-conditioned skill subset, count, and order; Agentic Compilation, SkillSmith, and SkCC compile workflows or skills into lower-cost, bounded, or security-checked runtime artifacts, while SkVM monitors outcomes across invocations, recompiles from accumulated failures, rolls back regressions, promotes stable code paths, and falls back to model execution \cite{wang2023voyager,li2026sgdr,wang2024awm,zabounidis2026scalar,zhao2026skillcomposer,chundru2026agenticcompilation,chen2026skvm,xu2026skillsmith,ouyang2026skcc}. SkillOps adds typed preconditions, artifacts, validators, and failure modes; filters skills by precondition; inserts validators or adapters; substitutes alternatives or repairs locally; and turns execution traces into persistent library updates \cite{song2026skillops}. GraSP closes the remaining composition candidate: it calibrates multi-skill retrieval confidence with historical success, routes low-confidence tasks to ReAct, compiles a typed precondition--effect DAG, verifies every node, applies bounded typed repairs, and escalates repair failure to global replanning or ReAct \cite{xia2026grasp}. Skill governance therefore reaches E3/D4 as a composed runtime chain in the frozen corpus. The open questions around GraSP concern external validity and transfer, not absence of the B5 control bundle. AIOS, AgentRM, and budget-aware value search govern execution resources \cite{mei2024aios,she2026agentrm,li2026bavt}. UALA, KnowNo, and Calibrate-Then-Act use uncertainty to select tools, request help, explore, or stop; Utility-Guided Agent Orchestration explicitly chooses respond, retrieve, tool call, verify, or stop from gain, cost, heuristic uncertainty, and redundancy \cite{han2024uala,ren2023knowno,ding2026cta,liu2026utility}.

\subsection{Implementation status and equivalence discipline}

We use four status descriptions. A \emph{released runtime interface} is an official, usable engineering boundary documented by its maintainer. A \emph{research prototype} implements and evaluates a mechanism in a bounded system. A \emph{partial mechanism} preserves only part of the historical state--control chain. A \emph{residual invariant} is the missing coupling left after the strongest current prototype is considered. These labels describe evidence and portability; repository popularity is excluded from scientific support.

Four equivalence tests prevent superficial mappings. A message bus transports content but lacks global-workspace semantics without capacity-limited competition and winner broadcast. A checkpoint persists state but lacks Soar semantics without a typed impasse and scoped substate. A handoff transfers work but lacks BDI semantics without a durable commitment and reconsideration rule. A guardrail constrains action but lacks MIDCA semantics without diagnosis and authority to change the active method. Section~\ref{sec:migration-depth} applies these tests to the strongest modern evidence.

%% file: figures/modern-runtime-clean.tex
\begin{figure}[H]
\centering
\begin{tikzpicture}[
  box/.style={draw=accentgreen!70!black, fill=accentgreen!8, rounded corners=2pt, align=center, text width=28mm, minimum height=10mm, font=\small},
  govern/.style={draw=accentorange!75!black, fill=accentorange!9, rounded corners=2pt, align=center, text width=55mm, minimum height=10mm, font=\small},
  flow/.style={-{Latex[length=2mm]}, thick},
  control/.style={-{Latex[length=2mm]}, thick, dashed},
  node distance=8mm and 10mm
]
\node[box] (state) {Authoritative state\\messages, goals, checkpoint};
\node[box, right=of state] (runner) {Runner / controller\\graph, actor, orchestrator};
\node[box, right=of runner] (policy) {Model policy\\reasoning and proposals};
\node[box, below=of policy] (tools) {Tools and sandbox\\approved execution};
\node[box, left=of tools] (events) {Observations / events\\results, errors, interrupts};
\node[box, left=of events] (persist) {Persistence update\\session, log, memory};
\node[govern, above=of runner] (governance) {Governance\\approvals, guardrails, budgets, stopping};
\draw[flow] (state) -- (runner);
\draw[flow] (runner) -- (policy);
\draw[flow] (policy) -- (tools);
\draw[flow] (tools) -- (events);
\draw[flow] (events) -- (persist);
\draw[flow] (persist) -- (state);
\draw[control] (governance) -- (runner);
\draw[control] (governance.east)
  -- ([xshift=7mm]tools.east |- governance.east)
  -- ([xshift=7mm]tools.east)
  -- (tools.east);
\node[draw, rounded corners=4pt, very thick, fit=(state)(policy)(tools)(persist)(governance), inner sep=6mm, label={[font=\small\bfseries]above:Runtime/framework boundary}] {};
\end{tikzpicture}
\caption{The modern unit of analysis. A policy such as ReAct or reflection runs inside a framework that owns authoritative state, transition control, execution, persistence, failure events, and governance. The cycle summarizes mechanisms documented across Letta, LangGraph, AutoGen, OpenHands, and the OpenAI Agents SDK \cite{packer2023memgpt,lettaRepo2026,langgraphRepo2026,autogenRepo2026,wang2025the,openaiAgentsSdkRepo2026}.}
\label{fig:runtime-levels}
\end{figure}
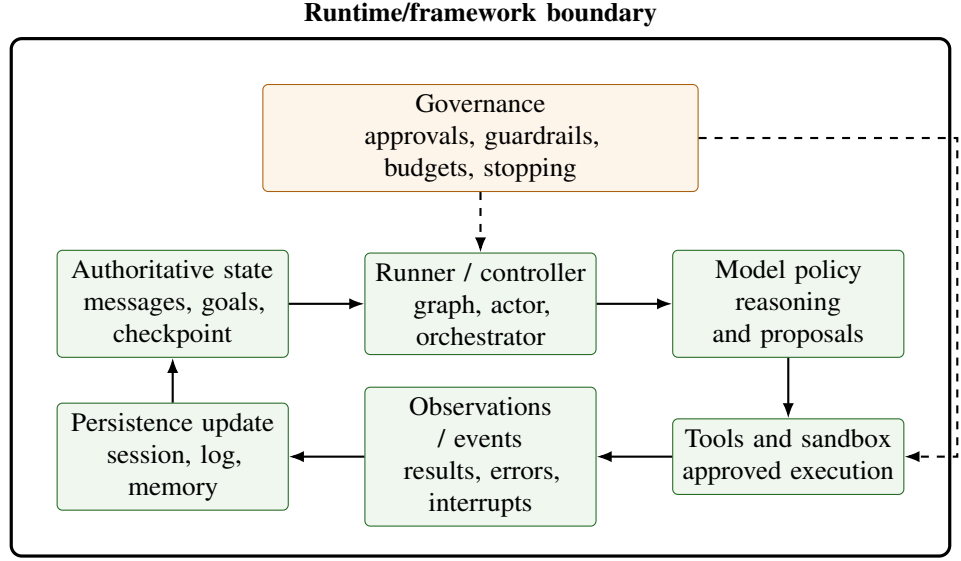

%% file: chapters/05_pairwise-structural-mapping.tex
\section{Migration Depth and Breakpoints}
\label{sec:migration-depth}

Across the frozen corpus, the result is an uneven migration landscape. State stores, tools, execution loops, checkpoints, message routing, memory operations, and workflow structures are widely available in the reviewed runtimes. Several adaptive control laws have also reached D3 or D4 in specialized systems. The residual value of cognitive architectures lies primarily in candidate control bundles whose individual edges have precedents but are not composed into a portable runtime invariant among the reviewed systems.

{\scriptsize
\begin{longtable}{@{}p{0.075\textwidth}p{0.17\textwidth}p{0.10\textwidth}p{0.075\textwidth}p{0.245\textwidth}p{0.255\textwidth}@{}}
\caption{Frozen-corpus migration ledger. D-levels apply to the named mechanism, not the complete architecture.}\label{tab:migration-ledger}\\
\toprule
Source & Highest-depth reviewed evidence & Status & E/D & Preserved control property & Residual breakpoint \\
\midrule
\endfirsthead
\toprule
Source & Highest-depth reviewed evidence & Status & E/D & Preserved control property & Residual breakpoint \\
\midrule
\endhead
\bottomrule
\endfoot
ACT-R & AgeMem; Memory-R1; MemCon; DeltaMem; A-MAC; CURATOR; DAM & Prototypes and design framework & E3/D4 & Outcomes or optimization update memory operations and admission; CURATOR couples online helpfulness to physical cost. & Activation-dependent accessibility, retrieval latency, and action/production utility are not one runtime law. \\
\midrule
Soar & PALADIN; AgentDebugX; Shepherd; AgentHER; Reflexion & Partial prototypes & E2/D3; E3/D3--D4 & Diagnosis changes recovery; execution can branch and revert; failure changes later behavior. & No reviewed system couples typed impasse, protected branch, typed return, and condition-safe compilation. \\
\midrule
LIDA & Global Workspace Agents; DyLAN; graph pruning & Research prototypes & E2/D3; E3/D4 & Limited state, selection, broadcast, write-back, and learned team/topology admission all have precedents. & Outcome-dependent admission of typed content coalitions is not integrated with the bounded broadcast cycle. \\
\midrule
Hearsay-II & AutoGen; AgentScope & Runtime interfaces & E4/D2 & Shared messages and state changes expose partial results to later components. & No common partial-solution value model or opportunistic agenda scheduler. \\
\midrule
BDI & Hybrid/self-aware BDI--LLM; CCA; Devil's Advocate & Deliberation and lifecycle prototypes & E1/D3 (BDI); E2/D3 (CCA); E3/D3 (plan revision) & Explicit BDI state, verified plans, intent graphs, deviation adjudication, and backtracking all exercise control. & No portable commitment lifecycle combines continuation, reconsideration, delegation, suspension, abandonment, and method authority. \\
\midrule
MIDCA & ADAS; AFlow; MetaReflection & Design-time prototypes & E3/D4 & Evaluation feedback changes methods, workflows, or reusable instructions across trials. & The reviewed systems do not give live diagnosis authority to replace the active planner, tool policy, scheduler, or stopping rule. \\
\midrule
CLARION & SCALAR & Research prototype & E2/D4 & Explicit specifications and learned policies update one another. & Cross-domain applicability, calibrated invocation, cheap fast path, and reliable fallback are not jointly guaranteed. \\
\midrule
ICARUS & SGDR; AWM; Voyager; SkillComposer; SkillOps; SkVM; GraSP; three static compilers & Research prototypes & E3/D4 (induction, governance, adaptive compilation, GraSP); E3/D3 (static compilers) & GraSP calibrates multi-skill routing, verifies typed DAG nodes, repairs locally, and escalates to replanning/ReAct. & B5 is closed in the corpus; remaining differences concern lineage, source fidelity, and cross-domain validity, not a missing runtime chain. \\
\midrule
EPIC & AIOS; AgentRM; budget-aware value search & Runtime and prototypes & E3/D3 & Budget and system state govern scheduling, admission, compaction, or search. & Token, time, tools, concurrency, memory, risk, and cognitive preemption remain separate policies. \\
\midrule
Sigma & UALA; KnowNo; Calibrate-Then-Act; Utility-Guided Orchestration & Research prototypes & E3/D3 & Explicit uncertainty and cost trigger tools, help, verification, exploration, or stopping. & One calibrated uncertainty state is not propagated through memory, planning, tools, execution, and stopping. \\
\end{longtable}
}

\input{figures/migration-depth-landscape}

\subsection{Four control-edge mappings}

Figure~\ref{fig:four-migration-cases} shows four cases where a shared feature label would be insufficient. Each mapping names the modern state object, trigger, authoritative edge, preserved invariant, and missing invariant.

\input{figures/migration-pairwise-cases}

\paragraph{ACT-R to adaptive memory control.}
ACT-R consumes buffer state and declarative activation; a retrieval or production match triggers selection; activation affects accessibility and latency, while utility affects action preference \cite{anderson2004integrated,actrmanual}. AgeMem, Memory-R1, MemCon, and DeltaMem learn memory operations or retrieval/consolidation policies from outcomes, while A-MAC optimizes a structured admission policy over future utility, confidence, novelty, recency, and content type \cite{yu2026agemem,yan2025memoryr1,jiang2026memcon,zhang2026deltamem,zhang2026amac}. CURATOR estimates online marginal helpfulness and retrieval propensity, subtracts harm, normalizes by bytes, and lets the resulting score govern retention, sharing, and trust \cite{wu2026curator}. DAM formalizes value and risk but contributes a design framework rather than an evaluated learning algorithm \cite{sun2025dam}. These systems support E3/D4 for experience-dependent memory control. B1 therefore makes no novelty claim for adaptive admission or value-cost memory governance; its missing invariant is the unified equation-level coupling among memory activation, retrieval time, downstream action utility, and maintenance cost.

\paragraph{Soar to typed recovery and experience learning.}
PALADIN classifies tool errors and retrieves recovery actions; AgentDebugX closes a Detect--Attribute--Recover--Rerun loop; Shepherd makes execution state first-class, reversible, forkable, and replayable; AgentHER and Reflexion change later behavior from failures \cite{vuddanti2025paladin,zhu2026agentdebugx,yu2026shepherd,ding2026agenther,shinn2023reflexion}. The reviewed systems therefore cover diagnosis, recovery, protected branching, and cross-episode learning in separate implementations. Soar's stronger chain couples a typed impasse to a protected substate, returns a typed resolution, and compiles only the dependencies that made the resolution valid \cite{soarmanual,soarchunking}. That coupling remains the B2 residual.

\paragraph{CLARION to explicit--learned skill coordination.}
SCALAR grounds explicit preconditions and effects into a learned policy and corrects the specification from trajectories, supporting E2/D4 \cite{zabounidis2026scalar}. SGDR, Agent Workflow Memory, and Voyager provide E3/D4 workflow induction and executable reuse, while SkillComposer learns task-conditioned skill subset, count, and order \cite{li2026sgdr,wang2024awm,wang2023voyager,zhao2026skillcomposer}. Agentic Compilation, SkillSmith, and SkCC establish static compile-and-execute fast paths, bounded runtime interfaces, and security-checked intermediate representations. SkVM adds capability-aware variants, cross-invocation outcome monitoring, failure-triggered recompilation, rollback, promotion, and code-path fallback, supporting E3/D4 for adaptive compilation \cite{chundru2026agenticcompilation,chen2026skvm,xu2026skillsmith,ouyang2026skcc}. SkillOps adds typed contracts, precondition filtering, validator and adapter insertion, alternative substitution, local repair, and trace-driven library updates, supporting E3/D4 \cite{song2026skillops}. GraSP composes the remaining edges: historical-success calibration governs multi-skill routing; typed DAG nodes expose pre/postconditions and verifiers; bounded repair preserves verified progress; and repair failure triggers global replanning or ReAct \cite{xia2026grasp}. The candidate B5 gap is therefore closed. Future work may test transfer beyond GraSP's evaluated domains, but that is an external-validity question rather than an unimplemented migration bundle.

\paragraph{EPIC/Sigma to resource and uncertainty control.}
AIOS gives scheduling and context resources an explicit kernel-level control surface, while AgentRM and budget-aware value search adapt admission or search to resource state \cite{mei2024aios,she2026agentrm,li2026bavt}. KnowNo, UALA, and Calibrate-Then-Act make uncertainty trigger help, tool use, exploration, or stopping; Utility-Guided Agent Orchestration couples heuristic uncertainty and step cost to respond, retrieve, tool, verify, and stop decisions \cite{ren2023knowno,han2024uala,ding2026cta,liu2026utility}. These are substantive D3 controls. EPIC contributes processor timing and cognitive bottlenecks; Sigma contributes a shared uncertainty state propagated through inference and decision \cite{kieras1997epic,sigma2011}. The residual is their coupling: calibrated uncertainty should persist across stages, allocate token, time, tool, and verification resources, and authorize interruption at a cognitive decision boundary.

\subsection{Translation analysis: where the semantics change}

\paragraph{Memory changes from accessibility law to policy action.}
ACT-R treats retrieval as a race among chunks whose activation determines both probability and latency, while production utility separately governs action choice. The modern memory systems make a different but substantial move: they expose store, retrieve, update, summarize, discard, or admit operations as policy actions and train or estimate their value \cite{yu2026agemem,sun2025dam,zhang2026deltamem,zhang2026amac}. The state object therefore changes from a chunk with an analytically defined activation history to a memory record interpreted by a learned controller. The trigger changes from a production request within a fixed cycle to an agent-selected memory operation. The learning scope also changes: modern systems update the policy that selects memory actions, while ACT-R updates activation and production utility in coupled but distinct subsystems. This explains both the D4 score and the residual. The useful migration target is not another storage API; it is a control law that lets predicted access time, memory value, and action value jointly determine whether information deserves scarce working context.

\paragraph{Failure changes from a control state to recovery content.}
Soar's impasse has architectural force before any recovery content is generated. The impasse type defines why the parent decision cannot continue; a substate creates a protected scope; a result crosses a typed boundary; chunking derives a future shortcut from the conditions that mattered \cite{soarmanual,soarchunking}. PALADIN already makes failure type causal, and AgentHER or Reflexion already makes failed experience reusable \cite{vuddanti2025paladin,ding2026agenther,shinn2023reflexion}. Their control ownership is distributed across an error classifier, recovery generator, memory or training process, and the host agent loop. This fragmentation matters because a generated recovery can mutate state before its diagnosis is validated, and a learned lesson may lack the conditions that make reuse safe. The remaining migration is an execution protocol that preserves parent state, constrains recovery authority, and records a resolution with explicit applicability conditions.

\paragraph{Workspace changes candidate type and broadcast semantics.}
DyLAN and adaptive graph pruning provide strong evidence that learned competition can reduce the active agent set or communication graph \cite{liu2023dylan,li2025agp}. Their candidates are agents and edges, so selection changes who participates. LIDA's candidates are typed content coalitions, so selection changes what every relevant process can use during a bounded cycle \cite{franklin2007lida}. The difference affects the preserved invariant. Both allocate scarce coordination capacity from contribution evidence, which justifies D4. Only the latter requires a shared winner whose provenance, urgency, confidence, and interference are exposed to memory, action, monitoring, and learning. A modern runtime can therefore inherit the competition law without inheriting the broadcast law. B3 tests whether content-level competition adds value after agent and topology selection are already optimized.

\paragraph{Commitment separates goal persistence from trajectory convergence.}
The hybrid BDI--LLM work demonstrates that an LLM can operate inside a controller where beliefs, goals, and plans remain explicit \cite{owayyed2025controlled}. Devil's Advocate shows that anticipatory reflection, post-action alignment, and backtracking can revise plan execution, but its prompt-centered control does not expose a durable intention object with authoritative continuation semantics \cite{wang2024devilsadvocate}. Goal-drift evaluation and premature-commitment diagnosis reveal opposite risks: the agent may abandon the intended objective, or it may converge too early on a trajectory that remains internally consistent but wrong \cite{arike2025goaldrift,mehta2026premature}. BDI contributes a normative control object: an intention is a commitment maintained under explicit success, impossibility, relevance, and reconsideration conditions \cite{rao1991bdi}. MIDCA adds a second authority boundary. A diagnosis should be able to change the reasoning method without silently changing the committed objective \cite{cox2016midca}. The combined migration target must therefore separate goal authority, method authority, and model proposal.

\paragraph{Skill migration reaches learning before it reaches governance.}
SCALAR is the strongest counterexample to a claim that explicit--implicit conversion is absent. Symbolic preconditions and effects guide policy grounding, and execution trajectories can revise the specification \cite{zabounidis2026scalar}. Voyager, SGDR, and Agent Workflow Memory show that experience can become reusable executable or state-grounded workflows; SkillComposer adds a learned policy over task-conditioned skill subsets and order \cite{wang2023voyager,li2026sgdr,wang2024awm,zhao2026skillcomposer}. D4 induction and composition edges are therefore present. SkVM further reaches D4 because outcomes across invocations trigger recompilation, rollback, code-path promotion, and fallback \cite{chen2026skvm}. SkillOps reaches D4 because typed validation and local repair feed persistent library maintenance \cite{song2026skillops}. GraSP then closes the composed selection-and-escalation chain with calibrated confidence, node verification, bounded repair, and global replanning/ReAct fallback \cite{xia2026grasp}. These decisions are architectural in GraSP rather than implicit in prompts: confidence controls routing, verification controls continuation, and repair exhaustion controls transfer to a global planner. These decisions determine whether proceduralization reduces model calls or creates a brittle shortcut. CLARION and ICARUS are useful here because they treat skill selection and representation form as control problems, not only as stored artifacts \cite{sun2006clarion,icarus2009}.

\paragraph{Resource and uncertainty controls remain locally owned.}
AIOS and AgentRM make scheduling, admission, context, and hibernation explicit system concerns; budget-aware value search changes search decisions as available budget changes \cite{mei2024aios,she2026agentrm,li2026bavt}. UALA, KnowNo, and Calibrate-Then-Act independently show that uncertainty can govern tools, help, exploration, and stopping \cite{han2024uala,ren2023knowno,ding2026cta,liu2026utility}. These systems establish that the control signals are actionable. Their state ownership is local to a scheduler, conformal set, tool decision, or search procedure. EPIC and Sigma suggest a stronger composition: processor demand and uncertainty should survive across stages so that one high-risk proposal can acquire verification budget, preempt a low-value action, and alter stopping criteria \cite{kieras1997epic,sigma2011}. The residual is a shared control plane, not an absence of scheduling or calibration.

\subsection{Why migration stalls at partial composition}

Three structural barriers recur across the mappings. First, control ownership is fragmented. Modern runtimes divide state, context, tools, evaluation, memory, and lifecycle hooks across independently designed interfaces \cite{meng2026agentharness}. A learned memory policy cannot preempt a tool scheduler unless both share a state model and authority protocol. A failure detector cannot compile a safe skill unless the runtime records the parent state, action dependencies, and postconditions. Historical architectures often specify these couplings centrally; software runtimes optimize replaceable components.

Second, model-generated representations are semantically unstable. Historical chunks, operators, intentions, and skill conditions have architecture-defined types. Language agents often store natural-language summaries, plans, or reflections whose meaning depends on a later model call. An explicit field therefore does not guarantee an executable invariant. Migration reaches D2 when the field exists, D3 when a trigger has authoritative consequences, and D4 when outcomes change that trigger or policy. The gap between D2 and D3 is frequently an authority gap rather than a representation gap.

Third, current evaluation rewards local capability more readily than cross-component invariants. A memory paper can measure retrieval and task success; a scheduler can measure throughput; an uncertainty method can measure calibration or help seeking. A composed runtime must measure causal state integrity, interruption, repeated failure, drift, transfer, and cost together. The five residual bundles are designed around this harder standard; the closed B5 case demonstrates that the standard can also eliminate an apparent gap. Each can fail because the historical invariant is ineffective, because the runtime does not grant it real authority, or because a simpler D3/D4 predecessor already captures the useful behavior.

\subsection{Remaining breakpoints outside the four cases}

Global Workspace Agents implements a limited working stage, attention selection, global broadcast, specialized proposal write-back, and an entropy-conditioned drive \cite{shang2026gwa}. It raises the LIDA-like workspace chain to E2/D3, while DyLAN and graph pruning reach E3/D4 for learned agent or topology admission \cite{liu2023dylan,li2025agp,franklin2007lida}. The narrower B3 residual is outcome-dependent admission learning over typed content coalitions inside the same bounded broadcast cycle. Hearsay-II remains relevant because AutoGen and AgentScope expose routing and shared-state boundaries without an opportunistic agenda over partial-solution value \cite{erman1980hearsay,autogenRepo2026,gao2025agentscope}.

Hybrid BDI--LLM control and the self-aware BDI agent implement explicit belief, desire, intention, plan, and deliberation structures, supporting E1/D3 in their domains \cite{owayyed2025controlled,lawton2026selfaware}. Cognitive Control Architecture adds an intent graph, deviation trigger, and tiered adjudicator for security-sensitive execution, while Devil's Advocate adds reflection-triggered backtracking \cite{liang2025cca,wang2024devilsadvocate}. These precedents narrow B4 to a portable commitment lifecycle with explicit continuation, reconsideration, delegation, suspension, abandonment, and authority to switch methods without silently changing the goal. MIDCA's D4 precedents still adapt workflows between trials rather than granting diagnosis immediate in-run method-switch authority \cite{hu2024adas,zhang2024aflow,cox2016midca}.

These cases explain why D4 cannot mean \emph{complete migration}. D4 states that a mapped control policy learns. Completeness requires the preserved and missing invariants to be assessed separately. Appendix~\ref{app:mapping-ledger} gives the expanded mapping ledger, and Appendix~\ref{app:recoding-packets} exposes every provisional D3/D4 claim for independent recoding.

%% file: figures/migration-depth-landscape.tex
\begin{figure}[t]
\centering
\begin{tikzpicture}[x=2.05cm,y=-0.56cm,font=\scriptsize]
\foreach \x/\lab in {0/D0,1/D1,2/D2,3/D3,4/D4}{
  \node[font=\scriptsize\bfseries] at (\x,0) {\lab};
  \draw[gray!28] (\x,0.35) -- (\x,10.45);
}
\foreach \y/\name in {1/ACT-R,2/Soar,3/LIDA,4/Hearsay-II,5/BDI,6/MIDCA,7/CLARION,8/ICARUS,9/EPIC,10/Sigma}{
  \node[anchor=east,font=\scriptsize\bfseries] at (-0.22,\y) {\name};
  \draw[gray!22] (0,\y) -- (4,\y);
}
\filldraw[fill=accentgreen!65,draw=accentgreen!70!black] (4,1) circle (2.2pt);
\filldraw[fill=accentgreen!65,draw=accentgreen!70!black] (4,2) circle (2.2pt);
\filldraw[fill=accentgreen!65,draw=accentgreen!70!black] (4,3) circle (2.2pt);
\filldraw[fill=accentblue!55,draw=accentblue!70!black] (2,4) circle (2.2pt);
\filldraw[fill=accentorange!70,draw=accentorange!80!black] (3,5) circle (2.2pt);
\filldraw[fill=white,draw=accentgreen!70!black,line width=0.8pt] (4,6) circle (2.5pt);
\node[anchor=south west,font=\tiny] at (4.05,6.02) {design-time};
\filldraw[fill=accentgreen!65,draw=accentgreen!70!black] (4,7) circle (2.2pt);
\filldraw[fill=accentgreen!65,draw=accentgreen!70!black] (4,8) circle (2.2pt);
\filldraw[fill=accentorange!70,draw=accentorange!80!black] (3,9) circle (2.2pt);
\filldraw[fill=accentorange!70,draw=accentorange!80!black] (3,10) circle (2.2pt);
\end{tikzpicture}
\par\smallskip
\parbox{0.92\textwidth}{\footnotesize D0: concept only; D1: interface; D2: explicit state/components; D3: runtime control law; D4: learning changes later control. Filled markers denote in-run or cross-episode control; the hollow MIDCA marker denotes design-time adaptation. Each marker is the highest-depth provisional code in the frozen corpus, not full historical fidelity.}
\caption{Provisional single-coder migration-depth landscape for all ten historical families in the frozen corpus. Most reach operational control or learning in specialized systems; Hearsay-II reaches explicit shared-state structure, while residual gaps concern source semantics and integration.}
\label{fig:migration-depth-landscape}
\end{figure}

%% file: figures/migration-pairwise-cases.tex
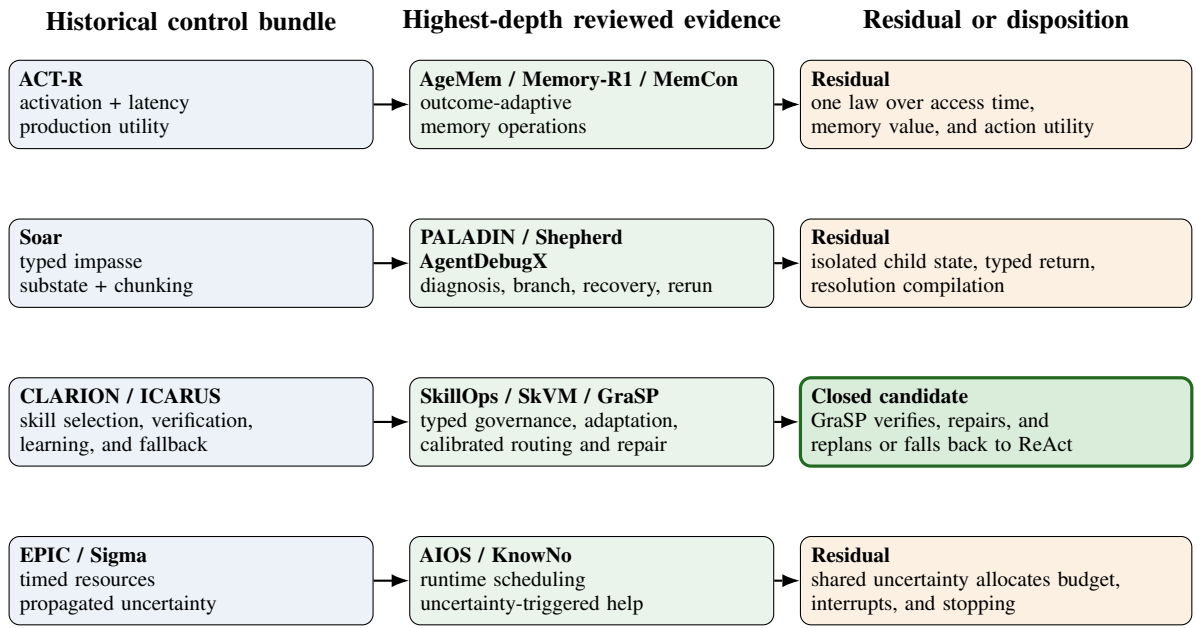
\begin{figure}[t]
\centering
\begin{tikzpicture}[
  font=\scriptsize,
  source/.style={draw,rounded corners,fill=accentblue!10,align=left,text width=0.25\textwidth,minimum height=0.92cm,inner sep=4pt},
  current/.style={draw,rounded corners,fill=accentgreen!10,align=left,text width=0.25\textwidth,minimum height=0.92cm,inner sep=4pt},
  residual/.style={draw,rounded corners,fill=accentorange!12,align=left,text width=0.27\textwidth,minimum height=0.92cm,inner sep=4pt},
  closed/.style={draw=accentgreen!70!black,very thick,rounded corners,fill=accentgreen!18,align=left,text width=0.27\textwidth,minimum height=0.92cm,inner sep=4pt},
  keep/.style={->,>=Latex,thick},
  miss/.style={->,>=Latex,thick,dashed}
]
\node[font=\small\bfseries] at (-5.3,4.25) {Historical control bundle};
\node[font=\small\bfseries] at (0,4.25) {Highest-depth reviewed evidence};
\node[font=\small\bfseries] at (5.35,4.25) {Residual or disposition};
\node[source] (a1) at (-5.3,3.15) {\textbf{ACT-R}\\activation + latency\\production utility};
\node[current] (a2) at (0,3.15) {\textbf{AgeMem / Memory-R1 / MemCon}\\outcome-adaptive\\memory operations};
\node[residual] (a3) at (5.35,3.15) {\textbf{Residual}\\one law over access time,\\memory value, and action utility};
\draw[keep] (a1) -- (a2); \draw[miss] (a2) -- (a3);
\node[source] (s1) at (-5.3,1.05) {\textbf{Soar}\\typed impasse\\substate + chunking};
\node[current] (s2) at (0,1.05) {\textbf{PALADIN / Shepherd}\\\textbf{AgentDebugX}\\diagnosis, branch, recovery, rerun};
\node[residual] (s3) at (5.35,1.05) {\textbf{Residual}\\isolated child state, typed return,\\resolution compilation};
\draw[keep] (s1) -- (s2); \draw[miss] (s2) -- (s3);
\node[source] (c1) at (-5.3,-1.05) {\textbf{CLARION / ICARUS}\\skill selection, verification,\\learning, and fallback};
\node[current] (c2) at (0,-1.05) {\textbf{SkillOps / SkVM / GraSP}\\typed governance, adaptation,\\calibrated routing and repair};
\node[closed] (c3) at (5.35,-1.05) {\textbf{Closed candidate}\\GraSP verifies, repairs, and\\replans or falls back to ReAct};
\draw[keep] (c1) -- (c2); \draw[keep] (c2) -- (c3);
\node[source] (r1) at (-5.3,-3.15) {\textbf{EPIC / Sigma}\\timed resources\\propagated uncertainty};
\node[current] (r2) at (0,-3.15) {\textbf{AIOS / KnowNo}\\runtime scheduling\\uncertainty-triggered help};
\node[residual] (r3) at (5.35,-3.15) {\textbf{Residual}\\shared uncertainty allocates budget,\\interrupts, and stopping};
\draw[keep] (r1) -- (r2); \draw[miss] (r2) -- (r3);
\end{tikzpicture}
\caption{Four deep mappings: three residual couplings and one closed candidate. Dashed arrows lead to surviving residuals; the solid green closure shows that GraSP composes calibrated routing, typed verification, bounded repair, and replanning/ReAct fallback \cite{anderson2004integrated,yu2026agemem,soarmanual,vuddanti2025paladin,sun2006clarion,icarus2009,song2026skillops,chen2026skvm,xia2026grasp,kieras1997epic,sigma2011,mei2024aios,ren2023knowno}.}
\label{fig:four-migration-cases}
\end{figure}

%% file: chapters/06_migration-agenda.tex
\section{Five Residual Control Bundles and One Closed Candidate}
\label{sec:residual-bundles}

The migration ledger initially produced six candidate control bundles. Adversarial closest-baseline screening closes B5 with GraSP, leaving five residual bundles. Each survivor is narrower than a historical architecture and stronger than a feature request: it specifies state, trigger, control authority, transition, and learning consequence that can be inserted into a current runtime and falsified against the highest-depth reviewed predecessor.

\subsection{A common runtime semantics}

The bundles can be expressed as control over a durable runtime state
\[
x_t=\langle i_t,w_t,m_t,s_t,u_t,b_t,q_t\rangle ,
\]
where $i_t$ is the active intention, $w_t$ bounded workspace content, $m_t$ memory state, $s_t$ compiled skills, $u_t$ uncertainty, $b_t$ resource budget, and $q_t$ the active failure or recovery substate. The language model produces a proposal or interpretation
\[
p_t=\operatorname{LLM}(w_t,i_t,q_t),
\]
while an external governor selects an authorized transition
\[
a_t=G(x_t,p_t),\qquad x_{t+1}=T(x_t,a_t,o_{t+1}).
\]
Learning changes later control only when an observed outcome $o_{t+1}$ updates the governor, admission policy, memory policy, or skill invocation policy. This distinction separates D3 from D4:
\[
\theta_{t+1}=L(\theta_t,x_t,a_t,o_{t+1}).
\]
Current runtimes already expose much of $x_t$ and $T$ through state, events, workflows, tools, checkpoints, or schedulers. They often leave $G$ implicit in application code or delegate it to another model call. The migration agenda makes the governor's state, trigger, authority, and learning edge explicit.

\subsection{From six candidates to five residual bundles}

The derivation begins with the coded atomic residual invariants in the mapping ledger. The merge rule requires a shared insertion boundary plus a dependency in which one atom supplies state, authority, or learning conditions for another. The split rule preserves separate bundles when triggers, baselines, or falsifiers remain independently testable. Table~\ref{tab:bundle-derivation} makes the resulting partition and its nearest rejected alternatives explicit.

{\scriptsize
\begin{longtable}{@{}p{0.07\textwidth}p{0.30\textwidth}p{0.29\textwidth}p{0.28\textwidth}@{}}
\caption{Derivation and closest-baseline disposition of six candidate bundles.}\label{tab:bundle-derivation}\\
\toprule
Bundle & Atomic residual invariants & Why merged & Closest alternative rejected \\
\midrule
\endfirsthead
\toprule
Bundle & Atomic residual invariants & Why merged & Closest alternative rejected \\
\midrule
\endhead
\bottomrule
\endfoot
B1 & Activation, retrieval latency, downstream utility, maintenance cost & All govern memory admission at context assembly; outcomes must update the same selector. & Kept separate from B3 because item accessibility and workspace capacity have different triggers and baselines. \\
\midrule
B2 & Typed diagnosis, protected substate, typed return, resolution compilation & Diagnosis defines recovery scope; return conditions define when a resolution is safe to compile. & Kept separate from B5 because recovery is an in-episode transition while general skill invocation is cross-episode. \\
\midrule
B3 & Outcome-dependent admission over typed coalitions inside a bounded broadcast cycle & Capacity allocation has value only when broadcast consequences update later content admission. & Not merged with B1 because candidates include tools, plans, monitors, and agents in addition to memory. \\
\midrule
B4 & Intention persistence, explicit reconsideration, live method-switch authority & A method switch needs a stable goal boundary, while reconsideration needs authority distinct from model proposals. & Goal and method control remain separately ablatable, but splitting them would omit the authority boundary that prevents method changes from rewriting intent. \\
\midrule
B5 & Calibrated multi-skill routing, typed verification, bounded repair, planner fallback & GraSP implements the complete chain and supplies component ablations. & Closed by closest-baseline screening; retained as a negative case, not a residual hypothesis. \\
\midrule
B6 & Propagated uncertainty, multi-resource allocation, preemption, interruption, stopping & Uncertainty becomes a control mechanism only when it changes resource or action authority across stages. & Resource scheduling and uncertainty can be ablated separately, but splitting them removes the causal allocation edge under test. \\
\end{longtable}
}

The initial six-candidate partition is reproducible conditional on the provisional source codes and stated rules; different source codes can change it. GraSP falsifies B5's absence condition, leaving B1--B4 and B6 as residual hypotheses. The closed case is methodologically important: the framework removes a research agenda item when a current system already composes its edges.

\subsection{Five residual hypotheses and one closed case}

\paragraph{B1: activation--latency--utility.}
A runtime memory item should carry activation, predicted retrieval latency, downstream value, and maintenance cost. Context assembly triggers competition under a fixed budget; selected items enter working context; observed use updates both accessibility and utility. Letta or LangGraph provides the insertion point; AgeMem, Memory-R1, MemCon, DeltaMem, A-MAC, and CURATOR are executable adaptive baselines; DAM is a conceptual comparator \cite{lettaRepo2026,langgraphRepo2026,yu2026agemem,yan2025memoryr1,jiang2026memcon,zhang2026deltamem,zhang2026amac,wu2026curator,sun2025dam}. Adaptive admission and value-cost memory governance are already implemented. CURATOR is the closest value-cost comparator; the hypothesis tests the remaining activation--latency--downstream-action-utility coupling.

\paragraph{B2: typed impasse--substate--compilation.}
An execution ledger should classify a failure, freeze the relevant parent state, open a bounded recovery subgraph, return a typed resolution, and compile a reusable recovery rule. LangGraph, OpenHands, Magentic-One, and Shepherd provide state and branching boundaries; PALADIN, AgentDebugX, AgentHER, and Reflexion provide diagnosis, recovery, and learning baselines \cite{langgraphRepo2026,wang2025the,fourney2024magentic,yu2026shepherd,vuddanti2025paladin,zhu2026agentdebugx,ding2026agenther,shinn2023reflexion}. The bundle fails under misdiagnosis, unsafe compilation, parent-state corruption, or no reduction in repeated failure.

\paragraph{B3: content competition--workspace--broadcast learning.}
Typed candidates from memory, agents, tools, monitors, and plans should compete for a bounded shared workspace. A winner becomes globally available for one cycle, and downstream utility updates later admission. Global Workspace Agents is the structural broadcast baseline; DyLAN and graph pruning are learned-admission baselines \cite{shang2026gwa,liu2023dylan,li2025agp}. The bundle fails if outcome-trained content admission adds no value over GWA's selection or simpler top-$k$ retrieval.

\paragraph{B4: intention--reconsideration--method authority.}
A persistent intention record should store the committed goal, rationale, plan, continuation condition, reconsideration triggers, authority, and termination status. A governor should decide whether to continue, delegate, suspend, abandon, or change method. Hybrid and self-aware BDI--LLM systems, Cognitive Control Architecture, Devil's Advocate, ADAS, and AFlow are the closest control baselines \cite{owayyed2025controlled,lawton2026selfaware,liang2025cca,wang2024devilsadvocate,hu2024adas,zhang2024aflow}. The bundle fails through stale commitment, oscillation, premature stopping, or method changes without diagnostic value.

\paragraph{B5: closed by GraSP.}
The predecessor chain already spans specification learning, induction, composition, typed contracts, local repair, adaptive compilation, rollback, and code-path fallback \cite{zabounidis2026scalar,li2026sgdr,wang2024awm,wang2023voyager,zhao2026skillcomposer,song2026skillops,chundru2026agenticcompilation,chen2026skvm,xu2026skillsmith,ouyang2026skcc}. GraSP composes the final selection--verification--repair--fallback protocol: calibrated confidence routes between a typed skill DAG and ReAct; every node has pre/postcondition checks; five typed operators perform bounded repair; and failed repair escalates to global replanning or ReAct \cite{xia2026grasp}. B5 is therefore an implemented E3/D4 convergence case. It is excluded from the residual intervention count and becomes the mandatory baseline for any proposed extension.

\paragraph{B6: uncertainty--resources--interruption--stopping.}
One calibrated uncertainty object should accompany each consequential proposal and propagate through memory admission, planning, tool authorization, execution monitoring, and stopping. AIOS, AgentRM, and the OpenAI Agents SDK expose scheduling and action-governance boundaries; UALA, KnowNo, Calibrate-Then-Act, and Utility-Guided Agent Orchestration provide the closest uncertainty and cost-sensitive action baselines \cite{mei2024aios,she2026agentrm,openaiAgentsSdkRepo2026,han2024uala,ren2023knowno,ding2026cta,liu2026utility}. Utility-Guided Orchestration selects respond, retrieve, tool, verify, or stop, but uses heuristic uncalibrated uncertainty and supplies no cross-stage propagation or interruption authority. The bundle fails under sequential miscalibration, excessive abstention, verification cost without risk reduction, or interruption that prevents correct completion.

{\scriptsize
\begin{longtable}{@{}p{0.12\textwidth}p{0.20\textwidth}p{0.25\textwidth}p{0.18\textwidth}p{0.19\textwidth}@{}}
\caption{Residual intervention agenda. Preregistration-ready protocol templates appear in Appendix~\ref{app:experiment-protocols}.}\label{tab:residual-agenda}\\
\toprule
Bundle & Runtime insertion point & Added control law & Closest baseline & Primary falsifier \\
\midrule
\endfirsthead
\toprule
Bundle & Runtime insertion point & Added control law & Closest baseline & Primary falsifier \\
\midrule
\endhead
\bottomrule
\endfoot
B1 memory & Context assembly & Activation, latency, value, and cost jointly select memory. & CURATOR; A-MAC; AgeMem; Memory-R1; MemCon; DeltaMem & Lower relevant recall or overhead without downstream use. \\
\midrule
B2 failure & Error/interrupt boundary & Typed substate recovers, returns, and compiles a resolution. & Shepherd; PALADIN; AgentDebugX; AgentHER & Misdiagnosis, state corruption, or no reuse. \\
\midrule
B3 workspace & Pre-inference admission & Outcome updates typed-content admission inside a bounded broadcast cycle. & GWA; DyLAN; graph pruning & No gain over GWA selection or top-$k$. \\
\midrule
B4 commitment & Workflow/governor boundary & Intention persists until explicit reconsideration; diagnosis may switch methods. & BDI--LLM; self-aware BDI; CCA; Devil's Advocate; ADAS; AFlow & Stale commitment, oscillation, or premature stopping. \\
\midrule

B6 resources & Scheduler and tool authorization & Shared uncertainty allocates budget, verification, interruption, and stopping. & Utility-Guided Orchestration; AIOS; AgentRM; CTA; KnowNo; UALA & Miscalibration or cost without risk reduction. \\
\end{longtable}
}

\subsection{Cross-bundle dependencies and implementation order}

\input{figures/control-bundle-dependencies}

The five residual bundles are not independent plug-ins, and the migrated B5 skill governor remains an implementation dependency. B4 supplies the commitment boundary that tells B2 which parent objective and state must remain stable during recovery. A successful B2 resolution can enter a GraSP-like skill graph only when failure type, dependencies, and postconditions define safe applicability. B1 and the implemented skill governor supply candidates to B3; B3 decides what enters the bounded inference workspace. B6 governs verification, tool authorization, interruption, and stopping across these controls.

This dependency structure suggests an implementation order. A runtime should first externalize intentions, uncertainty, budgets, and recovery state because later learning is unsafe when these authoritative objects remain hidden in prompts. It should then implement typed failure and bounded content admission, which create observable transition data. Adaptive memory and compiled skill policies can follow once the runtime can attribute outcomes to admitted evidence and authorized actions. Starting with automatic skill compilation before state and failure boundaries are explicit risks learning shortcuts whose conditions cannot be audited.

The dependency graph also defines compositional ablations. B2 can be tested with and without B4 to determine whether intention scope prevents recovery-induced goal drift. B3 can be tested with native retrieval, B1 candidates, and B1 plus B6 uncertainty to separate content value from risk-aware admission. A GraSP extension can admit every successful trace, only B2-verified resolutions, or no traces. These comparisons reveal whether value comes from the historical coupling or from a simpler component improvement.

\subsection{A bounded synthesis runtime}

\input{figures/runtime-synthesis-clean}

The synthesis runtime grants the language model proposal and interpretation authority while retaining durable control state outside the model call. State stores intentions, evidence, uncertainty, active substates, action history, and resource budgets. Memory and skills propose candidates; workspace admission limits what enters the next inference; the scheduler governs external actions; typed failure creates a recovery state; learning updates admission, memory, skills, and governance evidence. Each added edge has a current D3/D4 predecessor, so the research question is compositional value rather than rediscovery.

Evaluation must expose internal transitions as well as task outcomes. Matched model, prompt, tools, environment, and task budget are required. Success, latency, token/API cost, calibration, retries, interruption behavior, skill invocation, and state integrity should be measured together. A bundle is unsupported when a simpler current mechanism delivers the same outcome without its added state or authority.

%% file: figures/control-bundle-dependencies.tex
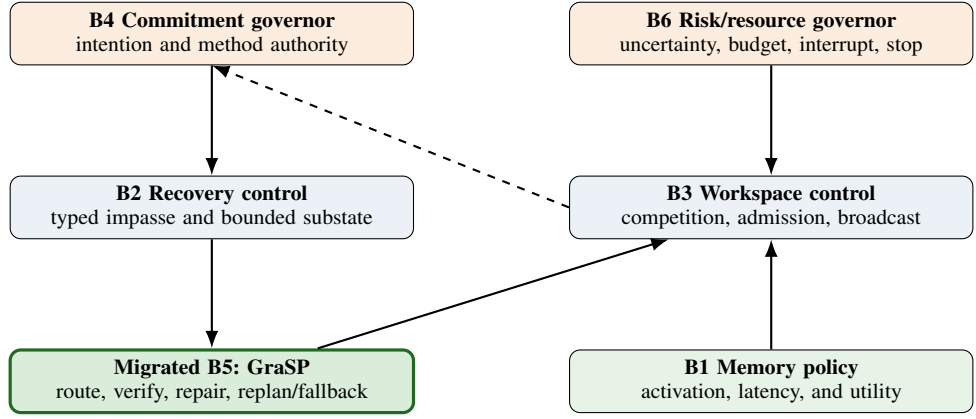
\begin{figure}[t]
\centering
\begin{tikzpicture}[
  font=\scriptsize,
  gov/.style={draw,rounded corners,fill=accentorange!14,align=center,text width=0.28\textwidth,minimum height=0.78cm},
  ctl/.style={draw,rounded corners,fill=accentblue!11,align=center,text width=0.28\textwidth,minimum height=0.78cm},
  migrated/.style={draw=accentgreen!70!black,very thick,rounded corners,fill=accentgreen!18,align=center,text width=0.28\textwidth,minimum height=0.78cm},
  asset/.style={draw,rounded corners,fill=accentgreen!11,align=center,text width=0.28\textwidth,minimum height=0.78cm},
  edge/.style={->,>=Latex,thick},
  feedback/.style={->,>=Latex,thick,dashed}
]
\node[gov] (b4) at (-3.7,2.3) {\textbf{B4 Commitment governor}\\intention and method authority};
\node[gov] (b6) at (3.7,2.3) {\textbf{B6 Risk/resource governor}\\uncertainty, budget, interrupt, stop};
\node[ctl] (b2) at (-3.7,0) {\textbf{B2 Recovery control}\\typed impasse and bounded substate};
\node[ctl] (b3) at (3.7,0) {\textbf{B3 Workspace control}\\competition, admission, broadcast};
\node[migrated] (b5) at (-3.7,-2.3) {\textbf{Migrated B5: GraSP}\\route, verify, repair, replan/fallback};
\node[asset] (b1) at (3.7,-2.3) {\textbf{B1 Memory policy}\\activation, latency, and utility};
\draw[edge] (b4) -- (b2);
\draw[edge] (b6) -- (b3);
\draw[edge] (b2) -- (b5);
\draw[edge] (b5) -- (b3);
\draw[edge] (b1) -- (b3);
\draw[feedback] (b3.west) -- (b4.south);
\end{tikzpicture}
\caption{Dependencies among five residual bundles and the migrated B5 skill governor. B4 scopes recovery; B6 governs risk and resources; B2 can produce verified resolutions for a GraSP-like governor; B1 and migrated skills propose assets to B3; workspace evidence can trigger B4 reconsideration.}
\label{fig:bundle-dependencies}
\end{figure}

%% file: figures/runtime-synthesis-clean.tex
\begin{figure}[t]
\centering
\begin{tikzpicture}[
  core/.style={draw=accentblue!70!black, fill=accentblue!8, rounded corners=2pt, align=center, text width=27mm, minimum height=10mm, font=\small},
  controlbox/.style={draw=accentgreen!70!black, fill=accentgreen!8, rounded corners=2pt, align=center, text width=29mm, minimum height=10mm, font=\small},
  wide/.style={draw=accentorange!75!black, fill=accentorange!8, rounded corners=2pt, align=center, text width=0.72\textwidth, minimum height=10mm, font=\small},
  flow/.style={-{Latex[length=2mm]}, thick},
  control/.style={-{Latex[length=2mm]}, thick, dashed}
]
\node[wide] (govern) at (0,2.55) {\textbf{Governance:} method choice, approvals, budgets, escalation, and stopping};
\node[core] (state) at (-4.8,0.8) {Persistent state\\intentions and substates};
\node[core] (llm) at (-1.6,0.8) {Bounded context\\and LLM proposal};
\node[core] (tools) at (1.6,0.8) {Scheduled tools\\and sandbox};
\node[core] (obs) at (4.8,0.8) {Observation / error\\and execution evidence};
\node[controlbox] (workspace) at (-4.8,-0.95) {Workspace selector\\activation and utility};
\node[controlbox] (uncertainty) at (-1.6,-0.95) {Uncertainty state\\inspect, ask, or act};
\node[controlbox] (scheduler) at (1.6,-0.95) {Resource scheduler\\priority and interrupt};
\node[controlbox] (failure) at (4.8,-0.95) {Failure diagnosis\\typed recovery state};
\node[wide] (learning) at (0,-2.7) {\textbf{Learning:} memory update, migrated skill governance, and evidence for later method selection};
\draw[flow] (state.east) -- (llm.west);
\draw[flow] (llm.east) -- (tools.west);
\draw[flow] (tools.east) -- (obs.west);
\draw[flow] (workspace.east) -- ++(4mm,0) |- (llm.south west);
\draw[flow] (uncertainty.north) -- (llm.south);
\draw[flow] (scheduler.north) -- (tools.south);
\draw[flow] (obs.south) -- (failure.north);
\draw[flow] (failure.south) |- (learning.east);
\draw[flow] (learning.west) -- ++(-5mm,0) |- (state.west);
\draw[control] (govern.south west) -- ++(0,-3mm) -| (llm.north);
\draw[control] (govern.south east) -- ++(0,-3mm) -| (scheduler.north east);
\draw[control] (uncertainty.east) -- (scheduler.west);
\node[draw, rounded corners=4pt, very thick, fit=(govern)(state)(obs)(learning), inner sep=5mm, label={[font=\small\bfseries]above:Proposed agent-runtime shell}] {};
\end{tikzpicture}
\caption{Synthesis architecture for testing the five residual integration hypotheses while reusing a migrated GraSP-like skill governor. Arrowheads attach to node ports and routes avoid label interiors. Solid arrows form execution and learning paths; dashed arrows denote control authority.}
\label{fig:migration-runtime-shell}
\end{figure}
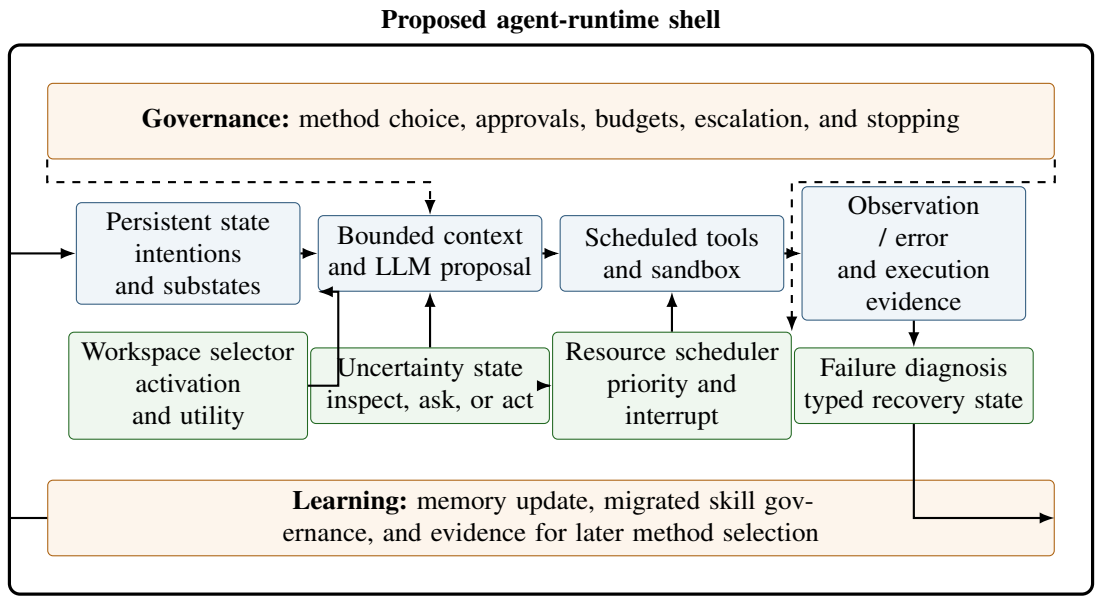

%% file: chapters/07_discussion-conclusion.tex
\section{Discussion and Limitations}
\label{sec:discussion}

\subsection{What the migration result changes}

The review changes the historical question from concept reuse to control-law depth. Memory, planning, reflection, scheduling, and uncertainty are already active research areas. The useful historical contribution is more specific: cognitive architectures state which object holds authority, which event triggers a transition, what remains stable during recovery, and how the outcome changes future control. Modern specialized systems demonstrate that many individual edges are feasible. General runtimes demonstrate that the required insertion points exist. The remaining challenge is to make these edges composable without importing every representational assumption of the source architecture.

This framing also changes the baseline standard. Adaptive memory must compare against AgeMem, Memory-R1, MemCon, DeltaMem, A-MAC, and CURATOR; failure control against Shepherd, PALADIN, AgentDebugX, AgentHER, and Reflexion; commitment and method authority against hybrid and self-aware BDI, CCA, Devil's Advocate, ADAS, and AFlow; compiled skills against GraSP, SCALAR, SGDR, Voyager, SkillOps, SkVM, SkillSmith, SkCC, and Agentic Compilation; and resource or uncertainty control against Utility-Guided Agent Orchestration, AIOS, AgentRM, UALA, KnowNo, and Calibrate-Then-Act. Historical inspiration has scientific value only when the residual edge survives these comparisons.

\subsection{Lineage, convergence, and engineering transfer}

E-level and D-level prevent two symmetric overclaims. Explicit historical lineage does not guarantee deep implementation, and deep functional implementation does not establish inheritance. CoALA and the hybrid BDI--LLM work provide explicit bridges \cite{sumers2023coala,owayyed2025controlled}; most other cases are structural or functional convergence. Engineering transfer should therefore cite the historical invariant being tested and the modern predecessor it extends. The claim is then traceable even when no genealogy exists.

The mechanism catalog is also not a demand for one universal cognitive architecture. Different applications may need different bundles. A short-lived tool call may not justify persistent intentions; a high-risk workflow may justify B4 and B6 while leaving B3 unnecessary. The proposed runtime is a test shell whose mechanisms should earn their cost through ablation, not a mandatory monolith.

\subsection{Threats to validity}

The corpus is bounded and rapidly changing. Its D-level records the highest-depth reviewed instance found by the cutoff date, so later systems may raise a row. ArXiv papers and official repositories differ in review status and stability. Repository evidence establishes an implementation surface, while mechanism and performance claims rely on the corresponding paper. The review does not infer prevalence from corpus size.

The mapping remains interpretive. The seven fields, claim packet, exclusion tests, and missing-invariant requirement make judgments inspectable, but they do not substitute for independent coding. The current manuscript does not claim inter-rater reliability. Appendix~\ref{app:recoding-packets} supplies a second-author packet for every D3/D4 result; scores must remain provisional until disagreements are recorded and adjudicated.

Historical architectures and modern agent systems also differ in purpose. Cognitive architectures seek stable accounts of cognition, while modern agents often optimize a task-specific composition of prompts, models, tools, and application code. Similar behavior can arise from different processes, and a mechanism that is useful for cognitive modeling may add unnecessary engineering overhead. Controlled interventions must therefore isolate the added state and control edge, hold the model and environment fixed, and report negative results.

Finally, this paper establishes feasibility and architectural hypotheses, not cross-domain causal value. Existing systems report local gains in their own settings. Those results do not prove that the five residual bundles improve reliability or efficiency when composed. GraSP closes B5 architecturally, but its reported results likewise do not establish universal transfer. The proposed experiments are valuable precisely because failure would narrow the role of the historical invariant.

%% file: chapters/08_conclusion.tex
\section{Conclusion}

Modern language agents have migrated much of the cognitive substrate, and adaptive control mechanisms have migrated to unequal depths. The strongest systems in the frozen corpus already learn memory operations, recover from typed failures, adapt teams and workflows, compile skills, schedule resources, and use uncertainty to govern action. Explicit residual-extraction and merge/split rules initially identify six candidate bundles. GraSP closes the skill-governance candidate by composing calibrated routing, typed verification, bounded repair, and global replanning/ReAct fallback. Five residual bundles remain: activation with latency and utility; impasse with scoped substate and compilation; workspace competition with broadcast learning; commitment with reconsideration and method authority; and uncertainty with resources, interruption, and stopping.

The paper contributes a catalog that explains what each historical mechanism does, an evidence--depth framework that separates lineage from implementation, and an intervention agenda that tests residual couplings against current D3/D4 baselines. This makes cognitive architectures useful as sources of precise runtime invariants and falsifiable control hypotheses for future agent development.

%% file: appendices/a_architecture-ledger.tex
\section{Complete Historical Architecture Ledger}
\label{app:architecture-ledger}

The main text explains the signature mechanism and purpose of each architecture. Tables~\ref{tab:architecture-sctp} and~\ref{tab:architecture-flr} retain the full seven-field extraction used in coding.

{\scriptsize
\begin{longtable}{@{}p{0.09\textwidth}p{0.24\textwidth}p{0.22\textwidth}p{0.20\textwidth}p{0.19\textwidth}@{}}
\caption{Historical state, control, trigger, and persistence fields.}\label{tab:architecture-sctp}\\
\toprule
Architecture & State substrate ($S$) & Control locus ($C$) & Trigger ($T$) & Persistence ($P$) \\
\midrule
\endfirsthead
\toprule
Architecture & State substrate ($S$) & Control locus ($C$) & Trigger ($T$) & Persistence ($P$) \\
\midrule
\endhead
\bottomrule
\endfoot
ACT-R & Typed buffers, chunks, procedural productions & Production matching and conflict resolution & Buffer match or module event & Chunks and learned utilities persist; buffers are transient \\
\midrule
Soar & Working memory, operators, preferences, productions & Operator decision cycle & Selection or typed impasse & Parent/substate working memory plus persistent chunks \\
\midrule
CLARION & Explicit rules and implicit networks/tendencies & Cross-level action integration & Context, rule match, or learned action evaluation & Rules and implicit skills persist \\
\midrule
LIDA & Perceptual content, coalitions, workspace, memories & Attention competition and action selection & Coalition formation and workspace competition & Memories persist; workspace is cycle-bounded \\
\midrule
Hearsay-II & Layered partial hypotheses on blackboard & Agenda scheduler over enabled knowledge sources & Blackboard change enables a source & Partial solution persists on blackboard \\
\midrule
BDI & Beliefs, desires, options, intentions, plans & Deliberation filter and intention executor & Belief change, option generation, reconsideration event & Intentions persist until release conditions \\
\midrule
MIDCA & Object-level state and meta-level trace & Object cycle plus meta-controller & Trace anomaly or failed expectation & Task state and diagnostic trace persist as configured \\
\midrule
ICARUS & Grounded concepts, goals, hierarchical skills & Goal-directed skill selection & Percept recognition and goal/skill applicability & Concepts and skill hierarchy persist \\
\midrule
EPIC & Perceptual, cognitive, and motor processor states & Cognitive processor under timing constraints & Perceptual event or processor completion & Task state persists; processor events are timed \\
\midrule
Sigma & Factor graph variables, factors, beliefs, decisions & Message passing and decision computation & New evidence or inference update & Graph structure and learned parameters persist \\
\end{longtable}

\begin{longtable}{@{}p{0.10\textwidth}p{0.25\textwidth}p{0.25\textwidth}p{0.25\textwidth}p{0.09\textwidth}@{}}
\caption{Historical failure, learning, resource, and source fields.}\label{tab:architecture-flr}\\
\toprule
Architecture & Failure semantics ($F$) & Learning operator ($L$) & Resource/governance ($R$) & Sources \\
\midrule
\endfirsthead
\toprule
Architecture & Failure semantics ($F$) & Learning operator ($L$) & Resource/governance ($R$) & Sources \\
\midrule
\endhead
\bottomrule
\endfoot
ACT-R & Retrieval failure, unavailable buffer, no applicable production & Activation and utility updates; procedural learning & Buffer capacity, module timing, central production bottleneck & \cite{anderson2004integrated,actrmanual} \\
\midrule
Soar & Typed tie, conflict, rejection, or no-change impasse & Chunking compiles substate resolution & Preferences and scoped problem spaces & \cite{laird1987soar,soarmanual,soarchunking} \\
\midrule
CLARION & Explicit/implicit inadequacy or conflict & Bottom-up extraction and top-down assimilation & Arbitration between costly explicit and fast implicit behavior & \cite{sun2006clarion} \\
\midrule
LIDA & Candidate loses competition or action fails & Broadcast-conditioned memory and action learning & Limited workspace capacity and competition & \cite{franklin2007lida} \\
\midrule
Hearsay-II & No enabled source or incompatible partial hypotheses & Primarily accumulation rather than a general adaptive operator & Agenda priority and blackboard levels & \cite{erman1980hearsay} \\
\midrule
BDI & Plan failure, impossible goal, or invalidated belief & Architecture-independent; commitment changes deliberation & Reconsideration and intention priority & \cite{rao1991bdi} \\
\midrule
MIDCA & Trace anomaly, expectation violation, method failure & Meta-level method adaptation & Meta-controller authority over object methods & \cite{cox2016midca} \\
\midrule
ICARUS & No applicable skill or failed subgoal/action & Skill acquisition and refinement & Hierarchical goal/skill applicability & \cite{icarus2009} \\
\midrule
EPIC & Delay, contention, or processor bottleneck & Learning is secondary in the architecture description & Explicit timing, parallelism, and bottlenecks & \cite{kieras1997epic} \\
\midrule
Sigma & Uncertain or inconsistent belief and low expected value & Parameter or structural updates in shared representation & Decision-theoretic control over uncertain state & \cite{sigma2011} \\
\end{longtable}
}

%% file: appendices/b_mapping-ledger.tex
\section{Expanded Mechanism Mapping Ledger}
\label{app:mapping-ledger}

{\scriptsize
\begin{longtable}{@{}p{0.09\textwidth}p{0.14\textwidth}p{0.16\textwidth}p{0.18\textwidth}p{0.20\textwidth}p{0.07\textwidth}@{}}
\caption{Expanded mapping claims and breakpoints.}\label{tab:expanded-mapping}\\
\toprule
Historical mechanism & Modern case & Matching state and trigger & Preserved control edge & Missing invariant & Code \\
\midrule
\endfirsthead
\toprule
Historical mechanism & Modern case & Matching state and trigger & Preserved control edge & Missing invariant & Code \\
\midrule
\endhead
\bottomrule
\endfoot
ACT-R activation/utility & AgeMem; Memory-R1; MemCon; DeltaMem; A-MAC; CURATOR; DAM & Memory item/value state; memory decision or admission event & Outcomes/optimization change later selection; CURATOR learns helpfulness and governs value per byte & No unified activation--latency--action utility law & E3/D4 \\
\midrule
Soar impasse/substate & PALADIN; AgentDebugX & Tool error or attributed cause; execution failure & Diagnosis chooses recovery and rerun & No protected child branch, typed return, or safe compilation & E3/D3 \\
\midrule
Soar protected substate & Shepherd & Reversible execution trace; fork/replay event & Past state becomes a protected transformable branch & No typed impasse, result return, or condition-safe compilation & E2/D3 \\
\midrule
Soar chunking & AgentHER; Reflexion & Failed trajectory or feedback after attempt & Experience changes later behavior & Learning is detached from a scoped impasse resolution & E3/D4 \\
\midrule
LIDA workspace & Global Workspace Agents & Limited working state; attention event & Selected content is broadcast and proposals write back & No outcome-trained content admission & E2/D3 \\
\midrule
LIDA learned competition & DyLAN; graph pruning & Agent/edge utility; task evidence & Learned admission changes active team/topology & Candidate type differs from content coalitions & E3/D4 \\
\midrule
Hearsay-II blackboard & AutoGen; AgentScope & Shared message/state events & State change enables later components & No common partial-solution semantics or opportunistic agenda law & E4/D2 \\
\midrule
BDI intentions & Hybrid and self-aware BDI--LLM & Belief/goal/plan state; deliberation event & Explicit BDI loop preserves structured commitment & Domain-specific, no portable reconsideration lifecycle & E1/D3 \\
\midrule
BDI plan revision & Devil's Advocate & Plan/subtask state; pre-action, post-action, completion reflection & Reflection can trigger alignment or backtracking & No durable authoritative intention and continuation semantics & E3/D3 \\
\midrule
BDI intent lifecycle & Cognitive Control Architecture & Intent graph; trajectory deviation event & Tiered adjudicator authorizes intervention & Security-specific; no complete commitment lifecycle or method authority & E2/D3 \\
\midrule
MIDCA meta-control & ADAS; AFlow; MetaReflection & Evaluation trace after trial & Feedback changes agent/workflow method & No live diagnosis with in-run method authority & E3/D4 \\
\midrule
CLARION explicit--implicit learning & SCALAR & Preconditions/effects and learned policy; execution trajectory & Specification and policy revise one another & No general confidence-governed fast path and fallback & E2/D4 \\
\midrule
ICARUS hierarchical skills & SGDR; AWM; Voyager; SkillComposer & State-grounded workflow or skill sequence; recognized task/context & Experience induces procedures and learned composition selects ordered skills & Calibrated cross-artifact applicability and safe fallback remain incomplete & E3/D4 \\
\midrule
ICARUS governed skill graph & SkillOps & Typed contract/graph and trace history; task match or failure & Preconditions filter selection; validators, alternatives, and repair govern execution; traces update the library & GraSP supplies the previously missing calibrated routing and planner fallback & E3/D4 \\
\midrule
ICARUS composed skill runtime & GraSP & Retrieved skills, confidence, typed DAG, verifier, repair budget; task or failure & Confidence routes control; pre/postconditions gate nodes; repair failure triggers replanning/ReAct & B5 closed; no historical lineage or universal transfer claim & E3/D4 \\
\midrule
ICARUS adaptive compiled path & SkVM & Compiled variant and outcome history; repeated failure or stable signature & Outcomes trigger recompilation, rollback, promotion, and code-path fallback & No calibrated cross-artifact selection or automatic return to full planning & E3/D4 \\
\midrule
ICARUS static compiled path & SkillSmith; SkCC; Agentic Compilation & Compiled artifact; matched task/runtime & Artifact bypasses repeated model reasoning & No outcome-updated applicability policy or full-planning fallback & E3/D3 \\
\midrule
EPIC resource timing & AIOS; AgentRM; BAVT & Budget, queue, and context state; admission/scheduling event & Resources change runtime selection & No unified cognitive preemption across resource and risk types & E3/D3 \\
\midrule
Sigma uncertainty control & UALA; KnowNo; CTA; Utility-Guided Orchestration & Uncertainty and cost; retrieve, tool, verify, or stop decision & Uncertainty directly changes action & Heuristic signal in the closest cost-sensitive controller; no durable calibrated cross-stage object or interruption & E3/D3 \\
\end{longtable}
}

The ledger deliberately allows multiple modern cases for one historical mechanism. Each correspondence has one E relation type; heterogeneous cases are split or receive case-specific labels. The strongest depth is reported only for the specific preserved edge, while the missing invariant prevents a D4 learning result from being interpreted as complete migration.

%% file: appendices/c_recoding-packets.tex
\section{D3/D4 Independent Recoding Packets}
\label{app:recoding-packets}

\begingroup
\scriptsize
\sloppy

The following packets are designed for independent second-author review. For each claim, the reviewer should inspect the cited historical source, modern primary paper, and official implementation where available; then assign E and D without seeing the provisional code. A disagreement record should name the failed necessary condition or exclusion test.

\subsection{R1: adaptive memory}
\textbf{Historical object:} ACT-R activation, latency, and utility \cite{anderson2004integrated,actrmanual}. \textbf{Modern evidence:} AgeMem, Memory-R1, MemCon, DeltaMem, and A-MAC implement outcome-adaptive or optimized memory control; CURATOR fits helpfulness online and uses retrieval propensity, value, harm, and byte cost to govern keep/share/trust; DAM is conceptual \cite{yu2026agemem,yan2025memoryr1,jiang2026memcon,zhang2026deltamem,zhang2026amac,wu2026curator,sun2025dam}. \textbf{State/trigger/edge:} memory item, learned helpfulness, retrieval propensity, and value-cost state; a memory-management or admission decision; outcome evidence updates future operations. \textbf{Preserved:} experience-dependent selection and value-cost governance. \textbf{Missing:} one activation--latency--action utility law. \textbf{Provisional code:} E3/D4.

\subsection{R2a: typed recovery}
\textbf{Historical object:} Soar typed impasse and substate \cite{soarmanual}. \textbf{Modern evidence:} PALADIN and AgentDebugX \cite{vuddanti2025paladin,zhu2026agentdebugx}. \textbf{State/trigger/edge:} execution error class or attributed root cause; failure; diagnosis selects recovery and rerun. \textbf{Preserved:} failure-specific control transition. \textbf{Missing:} parent isolation, substate decision cycle, typed result return, and safe compilation. \textbf{Provisional code:} E3/D3.

\subsection{R2b: reversible recovery branch}
\textbf{Historical object:} Soar protected substate \cite{soarmanual}. \textbf{Modern evidence:} Shepherd \cite{yu2026shepherd}. \textbf{State/trigger/edge:} first-class execution trace; meta-agent intervention; execution reverts, forks, edits, and replays from a prior state. \textbf{Preserved:} recovery can operate on a protected branch without destroying the original trace. \textbf{Missing:} typed impasse creation, typed return, and condition-safe compilation. \textbf{Provisional code:} E2/D3.

\subsection{R3: failure learning}
\textbf{Historical object:} Soar chunking \cite{soarchunking}. \textbf{Modern evidence:} AgentHER and Reflexion \cite{ding2026agenther,shinn2023reflexion}. \textbf{State/trigger/edge:} failed trajectory or feedback; episode completion; later behavior changes. \textbf{Preserved:} experience prevents some repeated failures. \textbf{Missing:} compilation from the dependency structure of a scoped impasse resolution. \textbf{Provisional code:} E3/D4.

\subsection{R4a: bounded workspace and broadcast}
\textbf{Historical object:} LIDA workspace competition and broadcast \cite{franklin2007lida}. \textbf{Modern evidence:} Global Workspace Agents \cite{shang2026gwa}. \textbf{State/trigger/edge:} limited working state; attention event; selected content broadcasts and specialized agents write proposals back. \textbf{Preserved:} selection has global control consequences inside a recurrent cycle. \textbf{Missing:} outcome-dependent learning of typed-content admission. \textbf{Provisional code:} E2/D3.

\subsection{R4b: learned competition}
\textbf{Historical object:} LIDA workspace admission \cite{franklin2007lida}. \textbf{Modern evidence:} DyLAN and adaptive graph pruning \cite{liu2023dylan,li2025agp}. \textbf{State/trigger/edge:} agent or edge contribution state; task evidence; active team/topology changes. \textbf{Preserved:} learned admission under limited computation. \textbf{Missing:} the learned candidate is an agent or edge rather than typed content in the broadcast cycle. \textbf{Provisional code:} E3/D4.

\subsection{R5a: persistent intention}
\textbf{Historical object:} BDI intention commitment \cite{rao1991bdi,kinny1991commitment,rao1995bdi}. \textbf{Modern evidence:} hybrid and self-aware BDI--LLM systems \cite{owayyed2025controlled,lawton2026selfaware}. \textbf{State/trigger/edge:} beliefs, goals, intentions, and plans; deliberation event; explicit BDI control selects and verifies execution. \textbf{Preserved:} structured commitment remains authoritative. \textbf{Missing:} portable cross-domain reconsideration lifecycle. \textbf{Provisional code:} E1/D3.

\subsection{R5b: reflection-triggered plan revision}
\textbf{Historical object:} BDI reconsideration policy \cite{kinny1991commitment,rao1995bdi}. \textbf{Modern evidence:} Devil's Advocate \cite{wang2024devilsadvocate}. \textbf{State/trigger/edge:} plan and subtask state; reflection event; the agent aligns, backtracks, or revises strategy. \textbf{Preserved:} execution evidence can change plan continuation. \textbf{Missing:} a durable authoritative intention object and explicit continuation, suspension, or abandonment rules. \textbf{Provisional code:} E3/D3.

\subsection{R5c: intent-graph adjudication}
\textbf{Historical object:} BDI commitment lifecycle \cite{kinny1991commitment,rao1995bdi}. \textbf{Modern evidence:} Cognitive Control Architecture \cite{liang2025cca}. \textbf{State/trigger/edge:} intent graph and action trajectory; deviation event; tiered adjudicator authorizes deeper control. \textbf{Preserved:} expected intent state constrains later action. \textbf{Missing:} domain-general continuation, delegation, suspension, abandonment, and method-switch authority. \textbf{Provisional code:} E2/D3.

\subsection{R6: method adaptation}
\textbf{Historical object:} MIDCA meta-control \cite{cox2016midca}. \textbf{Modern evidence:} ADAS, AFlow, and MetaReflection \cite{hu2024adas,zhang2024aflow,gupta2024metareflection}. \textbf{State/trigger/edge:} evaluation record; end of trial; agent or workflow method changes. \textbf{Preserved:} performance evidence changes the reasoning organization. \textbf{Missing:} live trace diagnosis and in-run method-switch authority. \textbf{Provisional code:} E3/D4 at design time.

\subsection{R7a: explicit--learned skill conversion}
\textbf{Historical object:} CLARION cross-level learning \cite{sun2006clarion}. \textbf{Modern evidence:} SCALAR \cite{zabounidis2026scalar}. \textbf{State/trigger/edge:} symbolic preconditions/effects and learned policy; skill execution trajectory; specification and policy revise one another. \textbf{Preserved:} bidirectional explicit--implicit adaptation. \textbf{Missing:} portable applicability confidence, cheap fast path, and safe fallback. \textbf{Provisional code:} E2/D4.

\subsection{R7b: induced reusable procedures}
\textbf{Historical object:} ICARUS hierarchical skills \cite{icarus2009}. \textbf{Modern evidence:} SGDR, Agent Workflow Memory, Voyager, and SkillComposer \cite{li2026sgdr,wang2024awm,wang2023voyager,zhao2026skillcomposer}. \textbf{State/trigger/edge:} state-grounded workflow, verified behavior, or task-conditioned skill sequence; recognized context or task; reusable procedure or ordered composition executes later. \textbf{Preserved:} experience creates reusable procedural control and learned selection. \textbf{Missing:} portable calibrated applicability and automatic fallback across artifacts. \textbf{Provisional code:} E3/D4.

\subsection{R7c: adaptive compiled path}
\textbf{Historical object:} ICARUS applicability-governed skills \cite{icarus2009}. \textbf{Modern evidence:} SkVM \cite{chen2026skvm}. \textbf{State/trigger/edge:} compiled variant, outcome history, failure logs, and code signature; repeated failure or stable signature; recompilation, rollback, promotion, or code-path fallback changes later execution. \textbf{Preserved:} experience updates the compiled fast path and can restore model execution. \textbf{Missing:} calibrated cross-artifact applicability, verified postconditions, and automatic return to full explicit planning. \textbf{Provisional code:} E3/D4.

\subsection{R7d: typed skill governance and maintenance}
\textbf{Historical object:} ICARUS applicability-governed skills \cite{icarus2009}. \textbf{Modern evidence:} SkillOps \cite{song2026skillops}. \textbf{State/trigger/edge:} typed contract graph and trace history; task matching or execution failure; precondition filtering, validators, adapters, alternatives, and repair govern execution, while traces update the library. \textbf{Preserved:} explicit applicability and outcomes causally change current and future procedural control. \textbf{Missing:} confidence calibrated to cross-artifact mis-invocation risk and automatic escalation to a full planner after local recovery exhaustion. \textbf{Provisional code:} E3/D4.

\subsection{R7e: composed skill governance}
\textbf{Historical object:} ICARUS applicability-governed skills \cite{icarus2009}. \textbf{Modern evidence:} GraSP \cite{xia2026grasp}. \textbf{State/trigger/edge:} retrieval confidence, typed DAG nodes, pre/postconditions, verifier, and repair budget; task arrival or node failure; confidence routes to DAG/ReAct, verification gates progress, typed repair preserves valid state, and failure escalates to replanning/ReAct. \textbf{Preserved:} applicability and execution evidence causally govern selection, continuation, repair, and fallback. \textbf{Missing:} no source-lineage claim or universal cross-domain validation. \textbf{Provisional code:} E3/D4. \textbf{Disposition:} closes candidate B5.

\subsection{R7f: static compiled fast path}
\textbf{Historical object:} ICARUS applicability-governed skills \cite{icarus2009}. \textbf{Modern evidence:} Agentic Compilation, SkillSmith, and SkCC \cite{chundru2026agenticcompilation,xu2026skillsmith,ouyang2026skcc}. \textbf{State/trigger/edge:} compiled artifact; matched task/runtime; the artifact executes with reduced model reasoning. \textbf{Preserved:} reusable explicit structure controls a cheaper path. \textbf{Missing:} outcome-updated applicability and full-planning escalation. \textbf{Provisional code:} E3/D3.

\subsection{R8: resource and uncertainty control}
\textbf{Historical object:} EPIC timing and Sigma uncertainty propagation \cite{kieras1997epic,sigma2011}. \textbf{Modern evidence:} AIOS, AgentRM, UALA, KnowNo, Calibrate-Then-Act, and Utility-Guided Agent Orchestration \cite{mei2024aios,she2026agentrm,han2024uala,ren2023knowno,ding2026cta,liu2026utility}. \textbf{State/trigger/edge:} budget, queue, context, uncertainty, and step cost; scheduling/help/retrieve/tool/verify/stop decision; control action changes execution. \textbf{Preserved:} resources and uncertainty are causal control state. \textbf{Missing:} calibrated cross-stage propagation plus cognitive preemption and interruption; the utility-guided signal is heuristic. \textbf{Provisional code:} E3/D3.

\endgroup

%% file: appendices/d_experiment-protocols.tex
\section{Expanded Experiment Protocol Templates}
\label{app:experiment-protocols}

Every intervention should hold the base model, prompt, tools, environment, and task budget fixed. Internal-state instrumentation is required because an outcome score alone cannot establish that the proposed control edge caused the difference. Before execution, each study must preregister the task population, experimental unit, model and dependency versions, stochastic replications, minimum sample size, metric formulas, equivalence or non-inferiority margin, uncertainty intervals or statistical tests, exclusion criteria, and stopping rule. In the table, a simpler baseline ``matches'' a bundle only when it is non-inferior under the preregistered primary margin while using no more resources. These templates define required decisions; task-specific values remain part of the future preregistration and are not reported as completed experiments.

{\scriptsize
\begin{longtable}{@{}p{0.11\textwidth}p{0.20\textwidth}p{0.23\textwidth}p{0.22\textwidth}p{0.18\textwidth}@{}}
\caption{Expanded measurement and rejection criteria.}\label{tab:expanded-experiments}\\
\toprule
Bundle & Required state trace & Primary outcomes & Safety/efficiency outcomes & Reject when \\
\midrule
\endfirsthead
\toprule
Bundle & Required state trace & Primary outcomes & Safety/efficiency outcomes & Reject when \\
\midrule
\endhead
\bottomrule
\endfoot
B1 & Activation, predicted latency, value, selected memories, downstream use & Task success, relevant recall, use precision & Token cost, stale-memory rate, selector latency & A simpler selector matches performance or relevant recall falls \\
\midrule
B2 & Failure type, parent hash, substate trace, returned resolution, compiled rule & Recovery by type, repeated-failure rate, transfer & Diagnosis latency, state corruption, unsafe reuse & Retry baseline matches recovery or compilation causes harm \\
\midrule
B3 & Candidate type, score, winner, broadcast recipients, admission update & Relevant-evidence density, distractor robustness & Context cost, rare-evidence suppression, turnover & Top-$k$ retrieval matches results or decisive evidence is suppressed \\
\midrule
B4 & Intention record, reconsideration event, diagnosis, method switch, termination & Goal drift, completion, correct stopping & Oscillation, replans, escalation precision, cost & Commitments become stale or method changes lack causal value \\
\midrule

B6 & Uncertainty provenance at each stage, allocated budget, interrupt and stop events & Selective risk, calibration, irreversible-action errors & Help precision, verification cost, interruption latency & Uncertainty fails to propagate or adds cost without reducing risk \\
\end{longtable}
}

At least three ablations are required: remove the new state object, remove the authoritative control edge while retaining logging, and remove the learning update while retaining the runtime transition. This separates representational overhead, immediate control value, and adaptive value.

%% file: appendices/e_implementation-boundaries.tex
\section{Implementation Evidence and Boundary Cases}
\label{app:implementation-boundaries}

\subsection{Versioned runtime evidence}

Repository records are used only to establish an executable interface and its inspected version. Scientific claims about mechanism behavior come from primary papers. Table~\ref{tab:versioned-runtimes} makes this separation explicit.

{\scriptsize
\begin{longtable}{@{}p{0.15\textwidth}p{0.19\textwidth}p{0.27\textwidth}p{0.20\textwidth}p{0.13\textwidth}@{}}
\caption{Version-pinned implementation evidence.}\label{tab:versioned-runtimes}\\
\toprule
Runtime & Official repository & Inspected revision & Interface used in mapping & Evidence role \\
\midrule
\endfirsthead
\toprule
Runtime & Official repository & Inspected revision & Interface used in mapping & Evidence role \\
\midrule
\endhead
\bottomrule
\endfoot
Letta & letta-ai/letta & b76da9092518 & Persistent agent state and memory-management boundary & Released runtime interface \cite{lettaRepo2026} \\
\midrule
LangGraph & langchain-ai/langgraph & 30c4d58db864 & State graph, checkpoint, interrupt, and resume boundary & Released runtime interface \cite{langgraphRepo2026} \\
\midrule
AutoGen & microsoft/autogen & 027ecf0a379b & Actor/event messaging and tool execution boundary & Released runtime interface \cite{autogenRepo2026} \\
\midrule
Microsoft Agent Framework & microsoft/agent-framework & c6442de52882 & Workflow, state, checkpoint, and approval boundary & Released runtime interface \cite{microsoftAgentFrameworkRepo2026} \\
\midrule
OpenAI Agents SDK & openai/openai-agents-python & c1b423749e2b & Run state, tools, handoffs, guardrails, approval, cancellation, and resume & Released runtime interface \cite{openaiAgentsSdkRepo2026} \\
\end{longtable}
}

The table uses twelve-character revision identifiers for readability; the bibliography records every full commit hash and access date.

AgentScope, OpenHands, Magentic-One, Agent Spec, and AIOS are analyzed primarily from their papers in this release \cite{gao2025agentscope,fourney2024magentic,wang2024openhands,wang2025the,amini2025open,mei2024aios}. Their official repositories support reproducibility where available, but no unreported moving-head state is used to raise an E-level or D-level.

\subsection{Boundary cases for E and D}

\paragraph{Checkpoint versus typed impasse.}
A checkpoint stores enough state to resume and therefore supports explicit persistence at D2. It reaches a Soar-related D3 claim only when a typed failure trigger opens a bounded child state with controlled return semantics. Generic resume remains a runtime substrate.

\paragraph{Message transport versus workspace broadcast.}
A message bus can expose a D1 interface and D2 message state. LIDA-like D3 control requires capacity-limited competition whose winner becomes broadly available for a cycle. Learned agent or topology pruning can be D4 for admission while still omitting content-level broadcast.

\paragraph{Handoff versus intention commitment.}
A handoff transfers authority to another component and can preserve task fields. BDI-like D3 control additionally requires a commitment that survives model proposals until an explicit continuation, delegation, suspension, abandonment, success, or impossibility condition fires.

\paragraph{Guardrail versus meta-control.}
A guardrail can block or transform an action at D3 when its trigger has runtime authority. MIDCA-like meta-control additionally diagnoses why the active method failed and chooses a replacement method. Blocking behavior alone does not establish metacognition.

\paragraph{Stored reflection versus learned control.}
A reflection record is D1 or D2 when it is merely available to a later prompt. It can support D4 for a narrow learning edge when later selection actually changes from feedback, as in Reflexion or MetaReflection \cite{shinn2023reflexion,gupta2024metareflection}. This does not imply Soar chunking or MIDCA method authority.

\paragraph{D4 versus complete migration.}
SCALAR reaches D4 because execution trajectories revise symbolic skill specifications and learned policies \cite{zabounidis2026scalar}. SkVM also reaches D4 because cross-invocation outcomes trigger recompilation, rollback, code-path promotion, and fallback \cite{chen2026skvm}. SkillOps supplies typed preconditions, validators, alternatives, local repair, and trace-driven library updates \cite{song2026skillops}. GraSP adds calibrated multi-skill routing, node-level pre/postcondition verification, bounded typed repair, and automatic global-replanning/ReAct fallback \cite{xia2026grasp}. Candidate B5 is therefore complete at E3/D4 within the frozen corpus; transfer beyond the evaluated settings remains an external-validity question.

%% file: appendices/f_full-evidence-ledger.tex
\section{Full Analytical Evidence Ledger}
\label{app:full-evidence-ledger}

The ledger records the analytical role of every architecture family, general runtime, explicit bridge, and mechanism-focused modern system in the frozen corpus. A record can support more than one row, but its status does not change across roles.

\subsection{Historical, runtime, and bridge records}

{\scriptsize
\begin{longtable}{@{}p{0.16\textwidth}p{0.19\textwidth}p{0.17\textwidth}p{0.27\textwidth}p{0.15\textwidth}@{}}
\caption{Historical architecture, runtime, and explicit-bridge evidence.}\label{tab:full-ledger-foundations}\\
\toprule
Record & Analytical role & Status & Mechanism evidence used & Source \\
\midrule
\endfirsthead
\toprule
Record & Analytical role & Status & Mechanism evidence used & Source \\
\midrule
\endhead
\bottomrule
\endfoot
ACT-R & Historical source & Architecture/manual & Buffers, activation, latency, production utility & \cite{anderson2004integrated,actrmanual} \\
\midrule
Soar & Historical source & Architecture/manual & Operators, typed impasses, substates, chunking & \cite{laird1987soar,soarmanual,soarchunking} \\
\midrule
CLARION & Historical source & Architecture theory & Explicit/implicit control and cross-level learning & \cite{sun2006clarion} \\
\midrule
LIDA & Historical source & Architecture theory & Coalition competition, global broadcast, learning & \cite{franklin2007lida} \\
\midrule
Hearsay-II & Historical source & Implemented architecture & Blackboard partial hypotheses and agenda scheduling & \cite{erman1980hearsay} \\
\midrule
BDI & Historical source & Formal architecture theory & Beliefs, options, intentions, commitment & \cite{rao1991bdi} \\
\midrule
MIDCA & Historical source & Architecture description & Object/meta cycle, trace diagnosis, method authority & \cite{cox2016midca} \\
\midrule
ICARUS & Historical source & Architecture description & Grounded concepts and hierarchical skills & \cite{icarus2009} \\
\midrule
EPIC & Historical source & Architecture theory & Processor timing, parallelism, bottlenecks & \cite{kieras1997epic} \\
\midrule
Sigma & Historical source & Architecture theory & Factor graph, uncertainty propagation, decision control & \cite{sigma2011} \\
\midrule
Letta/MemGPT & General runtime & Paper and released interface & Long-lived state and memory boundary & \cite{packer2023memgpt,lettaRepo2026} \\
\midrule
LangGraph & General runtime & Released interface & Stateful graph, checkpoint, interrupt, resume & \cite{langgraphRepo2026} \\
\midrule
AutoGen & General runtime & Paper and released interface & Event/actor messaging and orchestration & \cite{wu2023autogen,autogenRepo2026} \\
\midrule
AgentScope & General runtime & Primary paper & Agent, message, model, and tool interfaces & \cite{gao2025agentscope} \\
\midrule
OpenHands & General runtime & Primary papers & Event stream and agent--computer action boundary & \cite{wang2024openhands,wang2025the} \\
\midrule
Microsoft Agent Framework & General runtime & Released interface & Workflow, state, checkpoint, and approval & \cite{microsoftAgentFrameworkRepo2026} \\
\midrule
OpenAI Agents SDK & General runtime & Released interface & Run state, tools, handoff, approval, cancellation & \cite{openaiAgentsSdkRepo2026} \\
\midrule
AIOS & Runtime plus mechanism case & Research prototype & Scheduler, context, memory, tool, access managers & \cite{mei2024aios} \\
\midrule
Magentic-One & Supporting runtime & Primary paper & Orchestrator ledger and replanning & \cite{fourney2024magentic} \\
\midrule
Agent Spec & Supporting runtime & Primary paper & Declarative agent specification & \cite{amini2025open} \\
\midrule
Agentic software architecture review & Adjacent review & Review paper & BDI/deliberative models, typed tools, governance, and production architecture & \cite{alenezi2026evolution} \\
\midrule
CoALA & Explicit bridge & Conceptual architecture & Cognitive-science and symbolic-AI organization & \cite{sumers2023coala} \\
\midrule
LLM-ACTR & Explicit bridge & Hybrid study & LLM with ACT-R-like architecture control & \cite{cognitivellm2024} \\
\midrule
Bootstrapping Cognitive Agents & Explicit bridge & Hybrid study & LLM/cognitive-architecture combination & \cite{bootstrapping2024} \\
\midrule
Hybrid BDI--LLM & Explicit bridge and mechanism case & Research prototype & Rule BDI controller around LLM components & \cite{owayyed2025controlled} \\
\end{longtable}
}

\subsection{Mechanism-focused modern records}

{\fontsize{7.5pt}{7.9pt}\selectfont
\begin{longtable}{@{}p{0.16\textwidth}p{0.17\textwidth}p{0.18\textwidth}p{0.31\textwidth}p{0.12\textwidth}@{}}
\caption{Mechanism-focused evidence and its effect on the migration result.}\label{tab:full-ledger-modern}\\
\toprule
Record & Mechanism family & Status & Evidence contribution & Result effect \\
\midrule
\endfirsthead
\toprule
Record & Mechanism family & Status & Evidence contribution & Result effect \\
\midrule
\endhead
\bottomrule
\endfoot
AgeMem & ACT-R/memory & Research prototype & Memory operations become policy actions trained with reinforcement learning \cite{yu2026agemem}. & Raises memory control to D4 \\
\midrule
Memory-R1 & ACT-R/memory & Research prototype & PPO/GRPO train ADD, UPDATE, DELETE, and NOOP memory operations \cite{yan2025memoryr1}. & Independent D4 baseline \\
\midrule
MemCon & ACT-R/memory & Research prototype & Online contextual-bandit policy governs retrieval, plan injection, consolidation, and forgetting \cite{jiang2026memcon}. & Online D4 baseline \\
\midrule
A-MAC & ACT-R/memory & Research prototype & Cross-validated optimization learns structured admission over utility, confidence, novelty, recency, and content type \cite{zhang2026amac}. & Direct D4 admission baseline \\
\midrule
CURATOR & ACT-R/memory & Research prototype & Online helpfulness and retrieval propensity combine with harm and byte cost to govern keep, share, and trust \cite{wu2026curator}. & E3/D4 value-cost memory baseline \\
\midrule
DAM & ACT-R/memory & Conceptual design framework & Formalizes value, risk, and decision variables for memory management but contributes no evaluated control algorithm \cite{sun2025dam}. & Informs B1 design; does not raise D-level \\
\midrule
DeltaMem & ACT-R/memory & Research prototype & RL optimizes operation-level memory updating \cite{zhang2026deltamem}. & Adds independent D4 precedent \\
\midrule
PALADIN & Soar/failure & Research prototype & Tool-failure taxonomy selects execution-time recovery \cite{vuddanti2025paladin}. & Raises typed recovery to D3 \\
\midrule
AgentDebugX & Soar/failure & Research prototype & Detect--Attribute--Recover--Rerun couples root-cause diagnosis to repair and reusable diagnosis--repair bundles \cite{zhu2026agentdebugx}. & Strengthens integrated D3 recovery baseline \\
\midrule
Shepherd & Soar/failure & Research prototype & Reversible first-class traces support inspect, transform, fork, replay, and repair \cite{yu2026shepherd}. & E2/D3 protected-branch baseline \\
\midrule
AgentHER & Soar/failure & Research prototype & Failed trajectories become learning data \cite{ding2026agenther}. & Supports D4 across training \\
\midrule
Reflexion & Soar/failure & Research prototype & Verbal feedback changes later attempts \cite{shinn2023reflexion}. & D4 for narrow feedback edge \\
\midrule
DyLAN & LIDA/competition & Research prototype & Task evidence changes active agent team \cite{liu2023dylan}. & D4 for agent admission \\
\midrule
Adaptive graph pruning & LIDA/competition & Research prototype & Learned hard and soft pruning changes communication topology \cite{li2025agp}. & D4 for topology admission \\
\midrule
Global Workspace Agents & LIDA/workspace & Architecture prototype & Limited working state, attention selection, broadcast, and proposal write-back form a recurrent cycle \cite{shang2026gwa}. & E2/D3; narrows B3 to admission learning \\
\midrule
Hybrid BDI--LLM & BDI/commitment & Research prototype & Rule-based BDI retains structured control around LLMs \cite{owayyed2025controlled}. & E1/D3 precedent \\
\midrule
Self-aware BDI agent & BDI/commitment & Technical-report prototype & Implements explicit BDI deliberation, verification, generated plans, and simulation-seeded libraries \cite{lawton2026selfaware}. & E1/D3 commitment-loop baseline \\
\midrule
Devil's Advocate & BDI/commitment & Research prototype & Anticipatory reflection, post-action alignment, and backtracking revise plan execution \cite{wang2024devilsadvocate}. & E3/D3 plan-revision precedent \\
\midrule
Cognitive Control Architecture & BDI/commitment & Research prototype & Intent graph and deviation-triggered tiered adjudication govern security-sensitive execution \cite{liang2025cca}. & E2/D3 lifecycle baseline \\
\midrule
Goal-drift evaluation & BDI/commitment & Diagnostic study & Measures long-horizon deviation from intended goals \cite{arike2025goaldrift}. & Supports problem, not migration depth \\
\midrule
Premature commitment & BDI/commitment & Diagnostic study & Detects early trajectory convergence without correctness guarantee \cite{mehta2026premature}. & Defines opposite commitment risk \\
\midrule
ADAS & MIDCA/meta-control & Research prototype & Evaluation feedback searches code-represented agent designs \cite{hu2024adas}. & D4 at design time \\
\midrule
AFlow & MIDCA/meta-control & Research prototype & Evaluation feedback searches executable workflows \cite{zhang2024aflow}. & D4 at design time \\
\midrule
MetaReflection & MIDCA/meta-control & Research prototype & Past reflections update reusable instructions \cite{gupta2024metareflection}. & D4 across trials \\
\midrule
SGDR & ICARUS/skills & Research prototype & Online state-grounded workflow induction and retrieval \cite{li2026sgdr}. & D4 skill induction \\
\midrule
SCALAR & CLARION/skills & Research prototype & Symbolic specifications and learned policies revise one another \cite{zabounidis2026scalar}. & Strong E2/D4 case \\
\midrule
Voyager & ICARUS/skills & Research prototype & Verified behaviors compile into executable skills \cite{wang2023voyager}. & D4 procedural reuse \\
\midrule
Agent Workflow Memory & ICARUS/skills & Research prototype & Offline and online induction turns experience into selectively reused workflows \cite{wang2024awm}. & Direct D4 workflow-induction baseline \\
\midrule
SkillComposer & ICARUS/skills & Research prototype & Learned task-conditioned decoding selects skill subset, count, and order \cite{zhao2026skillcomposer}. & Direct D4 composition baseline \\
\midrule
SkillOps & ICARUS/skills & Research prototype & Typed contracts, validators, alternatives, local repair, and trace-driven maintenance govern current and future skill use \cite{song2026skillops}. & E3/D4 governance baseline \\
\midrule
GraSP & ICARUS/skills & Research prototype & Calibrated routing, typed DAG compilation, pre/post verification, bounded repair, and replanning/ReAct fallback form one runtime chain \cite{xia2026grasp}. & E3/D4 closes candidate B5 \\
\midrule
Agentic Compilation & ICARUS/skills & Research prototype & Deterministic workflow blueprint bypasses repeated model inference \cite{chundru2026agenticcompilation}. & E3/D3 fast-path baseline \\
\midrule
SkVM & ICARUS/skills & Research prototype & Cross-invocation outcomes trigger recompilation, rollback, stable-path promotion, and code-path fallback \cite{chen2026skvm}. & E3/D4 adaptive compiler/runtime baseline \\
\midrule
SkillSmith & ICARUS/skills & Research prototype & Boundary-first compilation produces minimal runtime interfaces \cite{xu2026skillsmith}. & E3/D3 bounded-interface baseline \\
\midrule
SkCC & ICARUS/skills & Research prototype & Typed intermediate representation supports cross-framework secure compilation \cite{ouyang2026skcc}. & E3/D3 portability/security baseline \\
\midrule
AIOS & EPIC/resources & Research prototype & Kernel managers schedule agents and govern context, memory, tools, and access \cite{mei2024aios}. & Resource control established \\
\midrule
AgentRM & EPIC/resources & Research prototype & Feedback governs admission, compaction, hibernation, and scheduling \cite{she2026agentrm}. & Adaptive resource precedent \\
\midrule
Budget-aware value search & EPIC/resources & Research prototype & Remaining budget changes value-tree selection \cite{li2026bavt}. & D3 budget control \\
\midrule
UALA & Sigma/uncertainty & Research prototype & Uncertainty arbitrates external-tool interaction \cite{han2024uala}. & D3 tool control \\
\midrule
KnowNo & Sigma/uncertainty & Research prototype & Calibrated uncertainty triggers help seeking \cite{ren2023knowno}. & D3 help control \\
\midrule
Calibrate-Then-Act & Sigma/uncertainty & Research prototype & Cost--uncertainty tradeoff changes exploration and stopping \cite{ding2026cta}. & D3 action/stopping control \\
\midrule
Utility-Guided Agent Orchestration & Sigma/uncertainty & Research prototype & Gain, cost, heuristic uncertainty, and redundancy select respond, retrieve, tool, verify, or stop \cite{liu2026utility}. & E3/D3 cost-sensitive action baseline \\
\end{longtable}
}

The diagnostic BDI records are intentionally separated from implementation evidence, and AIOS is cross-indexed as both runtime and mechanism case. This prevents problem evidence, released interfaces, and learned control laws from being counted as interchangeable support.

\subsection{Coverage, exclusions, and unsupported inferences}

The targeted audit used six documented query blocks. The search terms and the counterexamples that changed a mapping are summarized below; the dated research log retains replayable templates, exact identifiers, and record-level dispositions. Search-interface hit counts were not frozen.

{\scriptsize
\begin{longtable}{@{}p{0.12\textwidth}p{0.30\textwidth}p{0.27\textwidth}p{0.25\textwidth}@{}}
\caption{Targeted search blocks and residual effect.}\label{tab:search-blocks}\\
\toprule
Bundle & Query block after ``LLM agent'' & Material retained cases & Residual after screening \\
\midrule
\endfirsthead
\toprule
Bundle & Query block & Material retained cases & Residual after screening \\
\midrule
\endhead
\bottomrule
\endfoot
B1 & adaptive memory, memory policy, RL, consolidation, forgetting, admission & CURATOR; A-MAC; Memory-R1; MemCon; AgeMem; DeltaMem; DAM & activation--latency--action utility--cost coupling \\
\midrule
B2 & failure diagnosis, recovery, reversible trace, fork, replay, compilation & AgentDebugX; Shepherd; PALADIN & typed impasse-to-branch-to-safe-compilation coupling \\
\midrule
B3 & global workspace, limited capacity, broadcast, competition, admission & Global Workspace Agents; DyLAN; graph pruning & outcome-trained typed-content admission inside broadcast cycle \\
\midrule
B4 & BDI, intention, commitment, intent graph, reconsideration, method switch & BDI systems; CCA; Devil's Advocate; goal drift; premature commitment; ADAS; AFlow; MetaReflection & portable full commitment lifecycle plus method authority \\
\midrule
B5 & skill induction, composition, governance, compilation, fallback & GraSP; SkillOps; SkVM; AWM; SkillComposer; SGDR; SCALAR; Voyager; three static compilers & Closed by GraSP's calibrated routing, verification, repair, and replanning/ReAct fallback \\
\midrule
B6 & uncertainty, resource, budget, interrupt, stop, calibration & Utility-Guided Orchestration; AIOS; AgentRM; budget-aware value search; UALA; KnowNo; Calibrate-Then-Act & calibrated uncertainty propagated across all resource and action stages with interruption \\
\end{longtable}
}

The ledger is complete with respect to the frozen analytical corpus, not the entire cognitive-architecture or language-agent literature. Inclusion required an architecture-defining source, a primary modern paper exposing a control mechanism, a versioned official runtime interface, or an explicit historical--modern bridge. The review includes supporting runtime cases only when they locate a concrete insertion boundary used by a residual intervention. A record that supplies background terminology without changing a mechanism reconstruction, E-level, D-level, preserved invariant, or residual breakpoint remains outside the ledger.

Five categories were excluded from mechanism coding. Standalone retrieval augmentation was excluded because it does not by itself define an agent control cycle. Unversioned product pages were excluded because their implementation state cannot be reproduced. Capability demonstrations without authoritative state or transition exposure were excluded because output similarity cannot pass the node-and-edge claim test. Brain--Transformer analogies were excluded when they did not specify an agent operation. Performance-only reports were excluded from migration claims when the added component could not be isolated.

The corpus also records negative evidence carefully. Failure to find a direct lineage statement supports E3 or E4 wording only within the bounded search; it cannot prove that no influence occurred. Failure to find a complete control bundle identifies a residual invariant after the listed counterexamples; it cannot prove an empty research area. A diagnostic study can establish a control problem without raising migration depth. A released runtime can establish an insertion point without demonstrating that the proposed mechanism is effective.

Three forms of evidence remain outside the present manuscript. Independent second-author codes and disagreements have not yet been produced. No experiment compares the five residual bundles against their D3/D4 baselines. No systematic-search recall estimate establishes how much of the wider literature the bounded corpus captures. These absences are reported as validity limits rather than replaced with indirect evidence.